\documentclass[11pt,english,american]{article}
\usepackage[T1]{fontenc}
\usepackage[latin9]{inputenc}
\usepackage[letterpaper]{geometry}
\usepackage{array}
\usepackage{float}
\usepackage{multirow}
\usepackage{amsmath}
\usepackage{amssymb}
\usepackage{graphicx}
\usepackage{setspace}
\makeatletter

\providecommand{\tabularnewline}{\\}

\usepackage{scicite}

\usepackage{times}

\usepackage{etoolbox}
\patchcmd{\thebibliography}{\section*{\refname}}{}{}{}

\usepackage{caption}

\makeatother

\usepackage{babel}
\begin{document}
\title{\vspace{-1cm}Boltzmann Generators -- Sampling Equilibrium
States of Many-Body Systems with Deep Learning}
\date{$\:$}

\author{Frank No\'{e}$^{1,2,3,\dagger,*}$, Simon Olsson$^{1,\dagger}$,
Jonas K\"ohler$^{1,\dagger}$ and Hao Wu$^{4,1}$}

\maketitle
\vspace{-1cm}

$1$: FU Berlin, Department of Mathematics and Computer Science, Arnimallee
6, 14195 Berlin, Germany

$2$: FU Berlin, Department of Physics, Arnimallee 14, 14195 Berlin,
Germany

$3$: Rice University, Department of Chemistry, Houston, Texas 77005,
United States

$4$: Tongji University, School of Mathematical Sciences, Shanghai,
200092, P.R. China

$\dagger$: Equal contribution

{*}: Correspondence to: frank.noe@fu-berlin.de

\vspace{0.5cm}

\textbf{Abstract}: Computing equilibrium states in condensed-matter
many-body systems, such as solvated proteins, is a long-standing challenge.
Lacking methods for generating statistically independent equilibrium
samples in ``one shot'', vast computational effort is invested for
simulating these system in small steps, e.g., using Molecular Dynamics.
Combining deep learning and statistical mechanics, we here develop
Boltzmann Generators, that are shown to generate unbiased one-shot
equilibrium samples of representative condensed matter systems and
proteins. Boltzmann Generators use neural networks to learn a coordinate
transformation of the complex configurational equilibrium distribution
to a distribution that can be easily sampled. Accurate computation
of free energy differences and discovery of new configurations are
demonstrated, providing a statistical mechanics tool that can avoid
rare events during sampling without prior knowledge of reaction coordinates.

\clearpage

\section*{Introduction}

Statistical mechanics is concerned with computing the average behavior
of many copies of a physical system based on its microscopic constituents
and their interactions. For example, what is the average magnetization
in an Ising model of interacting magnetic spins, or what is the probability
of a protein to be folded as a function of the temperature? Under
a wide range of conditions, the equilibrium probability of a microscopic
configuration $\mathbf{x}$ (setting of all spins, positions of all
protein atoms, etc.) is proportional to $\mathrm{e}^{-u(\mathbf{x})}$,
for example, the well-known Boltzmann distribution. The dimensionless
energy $u(\mathbf{x})$ contains the potential energy of the system,
the temperature and optionally other thermodynamic quantities.

Except for simple model systems, we presently have no approach to
directly draw ``one-shot'', i.e., statistically independent, samples
$\mathbf{x}$ from Boltzmann-type distributions in order to compute
statistics of the system, such as free energy differences. Therefore,
one currently relies on trajectory methods, such as Markov-Chain Monte
Carlo (MCMC) or Molecular Dynamics (MD) simulations that make tiny
changes to $\mathbf{x}$ in each step. These methods sample from the
Boltzmann distribution in the long run, but many simulation steps
are needed to produce a statistically independent sample. This is
because complex systems often have metastable (long-lived) phases
or states and the transitions between them are rare events -- for
example, $10^{9}-10^{15}$ MD simulation steps are needed to fold
or unfold a protein. As a result, MCMC and MD methods are extremely
expensive and consume much of the worldwide supercomputing resources.

A common approach to enhance sampling is to speed up rare events by
biasing user-defined order parameters, or ``reaction coordinates'',
that may be of mechanical \cite{Torrie_JCompPhys23_187,Grubmueller_PhysRevE52_2893,LaioParrinello_PNAS99_12562,HeninEtAl_JCTC10_ABF},
thermodynamic \cite{SwendsenWang_PRL86_ParallelTempering,Hukushima_JPSJapan65_1604,MarinariParisi_EPL92_SimulatedTempering},
or alchemical nature \cite{Kirkwood_JCP35_CouplingParameterMethod,FrenkelSmit_MolecularSimulation}.
Applying these techniques to high-dimensional systems with a priori
unknown transition mechanisms is challenging, as identifying suitable
order parameters and avoiding rare events in other, unbiased directions,
becomes extremely difficult. For example, the development of enhanced
simulation protocols for the binding of small drug molecules to proteins
has become a research area in its own right \cite{KlimovichShirtsMobley_JCAMD15_FreeEnergy}.

Here we set out to develop a ``Boltzmann Generator'' machine that
is trained on a given energy function $u(\mathbf{x})$ and then produces
unbiased one-shot samples from $\mathrm{e}^{-u(\mathbf{x})}$, circumventing
the sampling problem without requiring any knowledge of reaction coordinates.
At first sight, this enterprise seems hopeless for condensed-matter
systems and complex polymers (Fig. \ref{fig:particle_dimer}a, Fig.
\ref{fig:bpti}b,c). In these systems, strongly repulsive particles
are densely packed, such that the number of low-energy configurations
are vanishingly few compared to the number of possible ways to place
particles.

Key to the solution is combining the strengths of deep machine learning
\cite{LeCunBengioHinton_DeepLearning_Nature05} and statistical mechanics
(Fig. \ref{fig:illustration}a): We train a deep invertible neural
network, to learn a coordinate transformation from $\mathbf{x}$ to
a so-called ``latent'' representation $\mathbf{z}$, in which the
low-energy configurations of different states are close to each other
and can be easily sampled, e.g. using a Gaussian normal distribution.
Enhancing MD sampling by user-defined\foreignlanguage{english}{ coordinate
transformations has been proposed previously} \cite{ZhuEtAl_PRL02_VariableTransformation}.
The novelty of Boltzmann Generators is that this transformation is
learned, and owing to the deep transformation network, can be as complicated
as needed to represent state changes in the many-body system. As Boltzmann
Generators are invertible, every sample $\mathbf{z}$ can be back-transformed
to a configuration $\mathbf{x}$ with high Boltzmann probability.
We can improve the ability to find relevant parts of configuration
space by ``learning from example'', where the potential energy $u(\mathbf{x})$
used to train the Boltzmann Generator is complemented by relevant
samples $\mathbf{x}$, e.g., from the folded or unfolded state of
a protein, but without knowing the probabilities of these states.
Then we employ statistical mechanics which offers a rich set of tools
to generate the target distribution $\mathrm{e}^{-u(\mathbf{x})}$
when the proposal distribution is sufficiently similar.

This paper demonstrates that Boltzmann Generators can be trained to
generate low-energy structures of condensed-matter systems and protein
molecules in one shot, as shown for model systems and a millisecond-timescale
conformational change of the BPTI protein. When the Boltzmann Generator
is initialized with a few structures from different metastable states,
it can generate statistically independent samples from these states
and can compute the free energy profiles of the corresponding transitions
without suffering from rare events. Although Boltzmann Generators
do not require reaction coordinates, they can be included in the training
in order to sample continuous free energy profiles and low-probability
states. When trained in this way, Boltzmann Generators can also generate
physically realistic transition pathways by performing simple linear
interpolations in latent space. We also show that multiple independent
Boltzmann Generators, trained on disconnected MD or MCMC simulations
of different states, can be employed to compute free energy differences
between these states in a direct and inexpensive way and without requiring
any reaction coordinates. Finally, we demonstrate that when employing
established sampling methods such as Metropolis Monte Carlo in the
latent space of a Boltzmann Generator, efficient methods can be constructed
to find new states and gradually explore state space.

\section*{Boltzmann Generators}

\begin{figure}[H]
\begin{centering}
\includegraphics[width=0.5\textwidth]{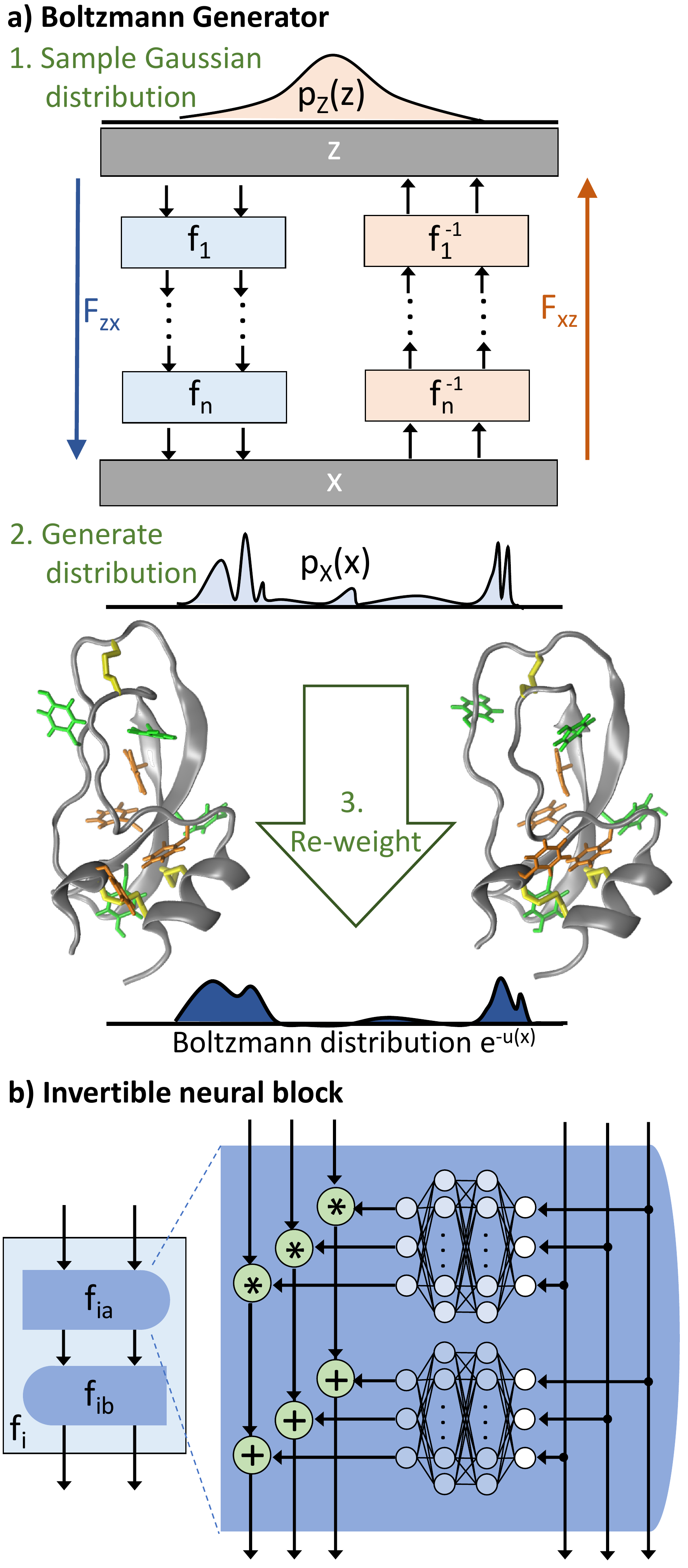}
\par\end{centering}
\caption{\label{fig:illustration}\textbf{Boltzmann Generators.} \textbf{a)}
A Boltzmann Generator is trained by minimizing the difference between
its generated distribution and the desired Boltzmann distribution.
Generation proceeds by drawing ``latent'' space samples $\mathbf{z}$
from a simple prior distribution (e.g., Gaussian) and transforming
them to configurations $\mathbf{x}$. The variable transformation
is formed by stacking invertible transformations $f_{1},...,f_{n}$
to a deep neural network $F_{zx}$ and its inverse, $F_{xz}$. To
compute thermodynamics, such as configurational free energies, the
samples must be reweighted to the Boltzmann distribution. \textbf{b)}
The Boltzmann Generator is composed of invertible neural network blocks.
Here, a non-volume-preserving transformation block is shown as an
example.}
\end{figure}

Neural networks that can draw statistically independent samples from
a desired distribution are called directed generative networks \cite{GoodfellowEtAl_GANs,KingmaWelling_ICLR14_VAE}.
Such networks have been demonstrated to generate photorealistic images
\cite{KarrasEtAl_ProgressiveGrowingGANs}, to produce deceivingly
realistic speech audio \cite{VanDenOord_WaveNet2}, and even to sample
formulae of chemical compounds with certain physico-chemical properties
\cite{GomezBombarelli_ACSCentral_AutomaticDesignVAE}. In these domains,
the exact target distribution is not known and the network is ``trained
by example'' using large databases of images, audio or molecules.
Here we are in the inverse situation, as we can compute the Boltzmann
weight of each generated sample $\mathbf{x}$, but we do not have
samples from the Boltzmann distribution a priori. The idea of Boltzmann
Generators is as follows (Fig. \ref{fig:illustration}a):
\begin{enumerate}
\item We learn a neural network transformation $F_{zx}$ such that when
sampling $\mathbf{z}$ from a simple prior, e.g., a Gaussian normal
distribution, $F_{zx}(\mathbf{z})$ will provide a configuration $\mathbf{x}$
which has a high Boltzmann weight, i.e. is coming from a distribution
$p_{X}(\mathbf{x})$ that is similar to the target Boltzmann distribution.
\item To obtain an unbiased sample and compute Boltzmann-weighted averages,
we reweight the generated distribution $p_{X}(\mathbf{x})$ to the
Boltzmann distribution $\mathrm{e}^{-u(\mathbf{x})}$. This can be
achieved with various algorithms; here the simplest one is used: assign
the statistical weight $w(\mathbf{x})=\mathrm{e}^{-u(\mathbf{x})}/p_{X}(\mathbf{x})$
to every sample $\mathbf{x}$ and then compute desired statistics,
such as free energy differences using this weight.
\end{enumerate}
\clearpage

\clearpage

For both, training and reweighting, it is important that we can compute
the probability $p_{X}(\mathbf{x})$ of generating a configuration
$\mathbf{x}$. This can be achieved when $F_{zx}$ is an invertible
transformation, which allows us to transform the known prior distribution
$p_{Z}(\mathbf{z})$ to $p_{X}(\mathbf{x})$ (Fig. \ref{fig:illustration}a,
Methods) \cite{TabakVandenEijnden_CMS10_DensityEstimation,DinhDruegerBengio_NICE2015}.
Physically, invertible transformations are analogous to flows of
a fluid that transform the probability density from configuration
space to latent space, or backwards. Volume-preserving transformations,
comparable to incompressible fluids were introduced in \cite{DinhDruegerBengio_NICE2015}.
Here we employ the non-volume preserving transformations introduced
in \cite{DinhBengio_RealNVP} (Fig. \ref{fig:illustration}b), as
they allow the probability distribution to be scaled differently at
different parts of configuration space. Alternatively, Boltzmann Generators
can be built using more general invertible transformations \cite{RezendeEtAl_NormalizingFlows,KingmaDhariwal_NIPS18_Glow,GrathwohlEtAl_FFJORD}.
Invertibility is achieved by adopting special neural network architectures
(Fig. \ref{fig:illustration}b; Methods). Multiple trainable invertible
``blocks'' can be stacked, thus encoding complicated variable transformations
in the form of a deep invertible neural network (Fig. \ref{fig:illustration}a).

Boltzmann Generators are trained by combining two modes: training
by energy and training by example. Training by energy is the main
principle behind Boltzmann Generators, and proceeds as follows: We
sample random vectors $\mathbf{z}$ from a Gaussian prior distribution,
and then transform them through the neural network to proposal configurations,
$\mathbf{x}=F_{zx}(\mathbf{z})$. In this way, the Boltzmann Generator
will generate configurations from a proposal distribution $p_{X}(\mathbf{x})$,
which, initially will be very different from the Boltzmann distribution,
and include configurations with very high energies. Next we compute
the difference between the generated distribution $p_{X}(\mathbf{x})$
from the Boltzmann distribution whose statistical weights $\mathrm{e}^{-u(\mathbf{x})}$
are known. For Boltzmann Generators, a natural measure of this difference
is the relative entropy, or Kullback-Leibler (KL) divergence. The
KL divergence can be computed as the following expectation value over
samples $\mathbf{z}$ (Methods):
\begin{equation}
J_{KL}=\mathbb{E}_{\mathbf{z}}\left[u(F_{zx}(\mathbf{z}))-\log R_{zx}(\mathbf{z})\right]\label{eq:main_loss_KL}
\end{equation}
Here, $u(F_{zx}(\mathbf{z}))$ is the energy of the generated configuration.
$R_{zx}$ is the determinant of the Boltzmann Generator's Jacobian
matrix, and measures how much the network scales the configuration
space volume at $\mathbf{z}$. The invertible network layers are designed
such that $R_{zx}$ can be easily computed (Methods). We treat $J_{KL}$
as a loss function: In order to train the Boltzmann Generator, we
approximate $J_{KL}$ using a batch of around 1000 samples, and then
change the neural network parameters so as to decrease $J_{KL}$.
A few hundred or thousand such iterations are required to train the
Boltzmann Generator for the examples in this paper. The resulting
few million computations of the potential energy in Eq. (\ref{eq:main_loss_KL})
are the main computational investment and take between one minute
and few hours for the present systems.

The KL divergence (\ref{eq:main_loss_KL}) is equivalent to the free
energy difference of transforming the Gaussian prior distribution
to the generated distribution (Methods, Supp. Mat.): The first term
$\mathbb{E}\left[u(F_{zx}(\mathbf{z}))\right]$ is the mean potential
energy, i.e. the internal energy of the system. The second term $\mathbb{E}\left[\log R_{zx}(\mathbf{z})\right]$
is equal to the entropic contribution to the free energy at the chosen
temperature, plus a constant factor. The terms in Eq. (\ref{eq:main_loss_KL})
counter-play in an interesting way: the first term tries to minimize
the energy, and therefore trains the Boltzmann Generator to sample
low-energy structures. The second term tries to maximize the entropy
of the generated distribution, and therefore prevents the Boltzmann
Generator from the so-called mode-collapse \cite{GoodfellowEtAl_GANs},
i.e. the repetitive sampling of a single minimum-energy configuration
which would minimize the first term.

Despite the entropy term in Eq. (\ref{eq:main_loss_KL}), training
by energy alone is not sufficient as it tends to focus sampling on
the most stable metastable state (Fig. \ref{fig_training_methods_RealNVP}).
We therefore additionally employ training by example, which is the
standard training method used in other machine learning applications,
and is here implemented with the maximum likelihood principle. We
initialize the Boltzmann Generator with some ``valid'' configurations
$\mathbf{x}$, e.g., from short initial MD simulations or an experimental
structure, transform them to latent space via $\mathbf{z}=F_{xz}(\mathbf{x})$.
Maximizing their likelihood in the Gaussian distribution corresponds
to minimizing the loss function \cite{TabakVandenEijnden_CMS10_DensityEstimation,DinhDruegerBengio_NICE2015}:
\begin{equation}
J_{ML}=\mathbb{E}_{\mathbf{x}}\left[\frac{1}{2}\left\Vert F_{xz}(\mathbf{x})\right\Vert ^{2}-\log R_{xz}(\mathbf{x})\right].\label{eq:main_loss_ML}
\end{equation}
Here, the first term $\frac{1}{2}\left\Vert F_{xz}(\mathbf{x})\right\Vert ^{2}$
is the energy of a harmonic oscillator corresponding to the Gaussian
prior distribution. Training by example is especially used in the
early stages of training, as it helps $F_{zx}$ to focus on relevant
parts of state space.

By combining training by energy and training by example, we can sample
configurations that have high probabilities and low free energies.
However, sometimes we want to sample states with low equilibrium probabilities,
such as transition states along a certain reaction coordinate (RC)
whose free energy profile is of interest. For this purpose, we introduce
an RC loss that can optionally be used to enhance the sampling of
a Boltzmann Generator along a chosen RC (Methods).

\section*{Results}

\subsubsection*{Illustration on model systems}

We first illustrate Boltzmann Generators using two-dimensional model
potentials that have metastable states separated by high energy barriers:
the double well potential, and the Mueller potential (Fig. \ref{fig:model_systems}a,g).
MD simulations stay in one metastable state for a long time before
a rare transition event occurs. Hence, the distributions in configuration
space $(x_{1},x_{2})$ are split into two modes (Fig. \ref{fig:model_systems}a,g,
transition state and intermediate state ensembles are shown in yellow
for clarity but are not used for training). We are training Boltzmann
Generators using the two short and disconnected simulations whose
samples are shown in Fig. \ref{fig:model_systems}a,g (details in
Supp. Mat., convergence in Fig. \ref{fig_convergence1}). Fig. \ref{fig:model_systems}b,h
show the latent spaces learned by the Boltzmann Generator, note that
their exact appearance varies between different runs due to stochasticity
in neural network training. In both latent spaces, the probability
densities of the two states and the transition/intermediate states
are ``repacked'' so as to form a density concentrated around the
origin. 

We use the Boltzmann generators by sampling from their latent spaces
according to Gaussian distributions. After transforming these variables
via $F_{zx}$, this produces uncorrelated and low-energy samples from
both stable states without any sampling problem (Fig. \ref{fig:model_systems}c-d,i-j)
. A variety of training methods succeed in sampling across the barrier
such that the rare event nature of the system is eliminated (Fig.
\ref{fig_training_methods_RealNVP}). Using a Boltzmann Generator
trained by energy and by example with simple reweighting reproduces
the precise free energy differences of the two metastable states,
although no reaction coordinate is employed to indicate the direction
of the rare event (Fig. \ref{fig:model_systems}e,k; green). By additionally
training with the RC loss to promote sampling along $x_{1}$ (double
well) or $x_{2}$ (Mueller potential), the low-probability transition
states are sampled (Fig. \ref{fig:model_systems}d,j;, orange), and
the full free energy profile can be reconstructed with high precision
(Fig. \ref{fig:model_systems}e,k; orange).

The Boltzmann Generator repacks the high-probability regions of configuration
space into a concentrated latent space density. We therefore ask about
the physical interpretation of direct paths in latent space. Specifically,
we interpolate linearly between the latent space representations of
samples from different energy minima, similar as it is done with generative
networks in other disciplines \cite{KingmaDhariwal_NIPS18_Glow,GomezBombarelli_ACSCentral_AutomaticDesignVAE}.
When mapping these linear interpolations back to configuration space,
they result in nonlinear pathways that have low-energies and high
probabilities (Fig. \ref{fig:model_systems}f,l). Although there is
no general guarantee that linear paths in latent space will result
in low energies, this result indicates that the latent spaces learned
by Boltzmann Generators can be used to provide candidates of order
parameters for bias-enhanced or path-based sampling methods \cite{Torrie_JCompPhys23_187,LaioParrinello_PNAS99_12562,BolhuisChandlerDellagoGeissler_AnnuRevPhysChem02_TPS}.

For the double-well system, the unbiased MD simulation needs on average
$4\cdot10^{6}$ MD steps for a single return trip between the two
states (Supp. Mat.), and about $100$ such crossings are required
to compute the free energy difference with the same precision as the
Boltzmann Generator results shown in (Fig. \ref{fig:model_systems}e).
The total effort of training the Boltzmann Generator (including generating
the initial simulation data) corresponds to about $10^{6}$ steps,
but once this is done, statistically independent samples can be generated
at no significant cost. For this simple system, the Boltzmann Generator
is therefore about a factor $100$ more efficient than direct simulation,
but much more extreme savings can be observed for complex systems,
as shown below.
\begin{center}
\begin{figure}[H]
\begin{centering}
\includegraphics[width=1\textwidth]{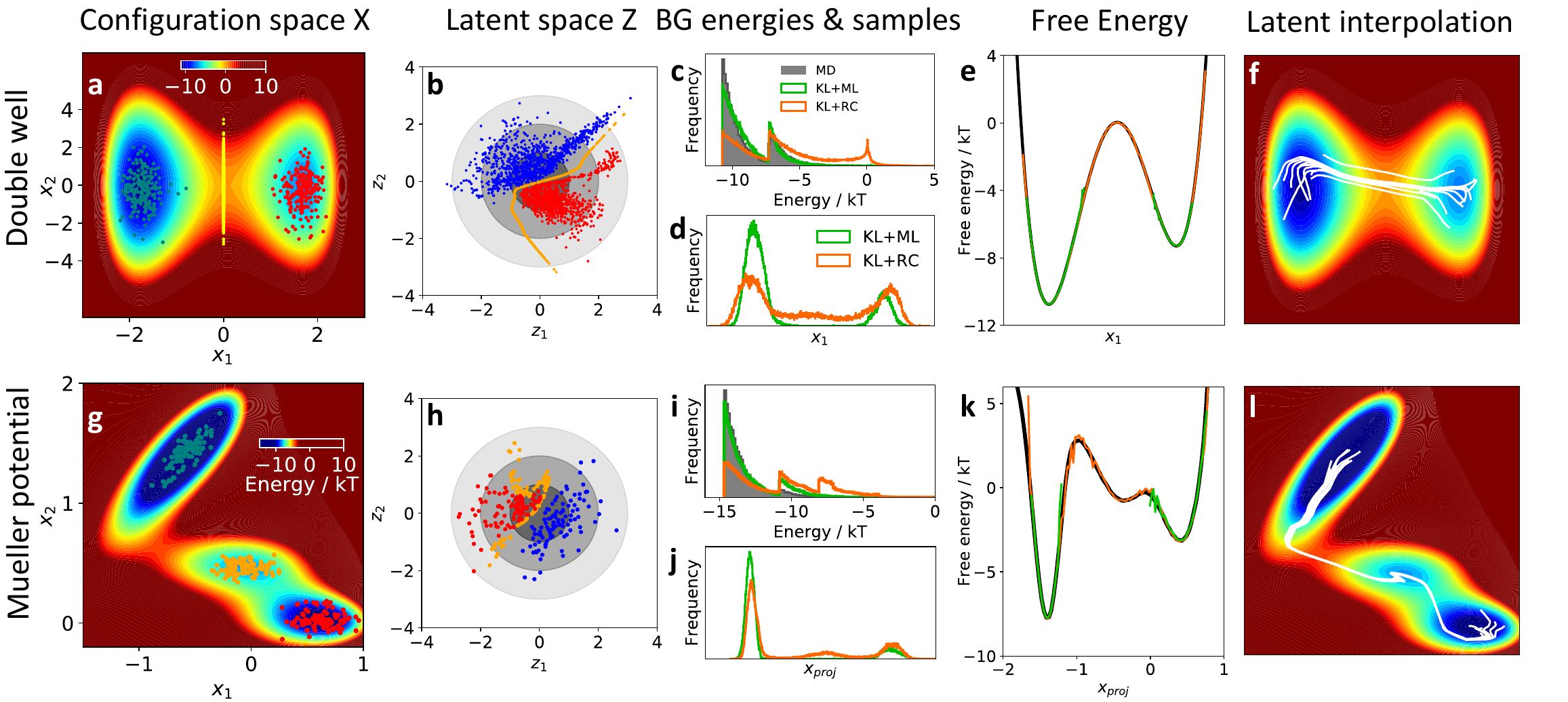}
\par\end{centering}
\caption{\label{fig:model_systems}\textbf{Application of Boltzmann Generators
on two-dimensional bistable systems}. \textbf{a},\textbf{g)} Two-dimensional
potentials: double well ($x_{1}$ is the slow coordinate) and Mueller
potential. Two short MD simulation trajectories (blue, red) stay in
their metastable states without crossing. Transition state and intermediate
state ensembles are shown (orange) but not used for training.\textbf{
b},\textbf{h)} Latent-space distribution of trajectories shown in
a,g) when mapped through trained $F_{xz}$. \textbf{c},\textbf{i})
potential energy distribution sampled by MD simulation (grey) and
by Boltzmann Generators trained by energy and by example (KL+ML, green)
and using reaction coordinate training (KL+RC, orange). \textbf{d},\textbf{j})
Boltzmann Generator sample distribution along the slow coordinates.
For the Mueller potential, $x_{\mathrm{proj}}$ is defined as projection
along the vector $(1,-1)$.\textbf{ e},\textbf{k)} Free energy estimates
obtained from Boltzmann Generator samples after reweighting. \textbf{f},\textbf{l)}
Paths generated by linear interpolation in Boltzmann Generator latent
space (b,h) between random pairs of ``blue'' and ``red'' MD samples.}
\end{figure}
\par\end{center}

\subsubsection*{Thermodynamics of condensed-matter systems}

As a second example, we demonstrate that Boltzmann Generators can
sample high-probability structures and efficiently compute the thermodynamics
in crowded condensed matter systems. We simulated a dense system of
two-dimensional particles confined to a box as suggested in \cite{NilmeyerEtAl_PNA11_NCMC}
(Fig. \ref{fig:particle_dimer}a). Immersed in the fluid is a bistable
particle dimer whose open and closed states are separated by a high
barrier (Fig. \ref{fig:particle_dimer}b). Opening or closing the
dimer directly is not possible due to the high density of the system,
but requires concerted rearrangement of solvent particles. At close
distances, particles repel each other strongly, resulting in a crowded
system. Thus, the fraction of low-energy configurations is vanishingly
small, and manually designing a sampling method that simultaneously
places all $38$ particles and achieves low energies appears unfeasible.

We train a Boltzmann Generator to sample one-shot low-energy configurations
and use it in order to compute the free energy profiles of opening
/ closing the dimer. Key to treat explicit-solvent systems such as
this one is to incorporate the permutation invariance of solvent molecules.
If physically identical solvent molecules would be distinguished,
every exchange of solvent molecule positions due to diffusion would
represent a new configuration, resulting in an enormous configuration
space even for this 38-particle system. We therefore remove identical-particle
permutations from all configurations input into or sampled by the
Boltzmann Generator by exchanging particle labels so as to minimize
the distance to a reference configuration (Supp. Mat.).

The training is initialized with examples from separate, disconnected
simulations of the open and closed states, but in later stages, training
by energy (\ref{eq:main_loss_KL}) dominates (Supp. Mat., Fig. \ref{fig_convergence1},
Table \ref{tab:hyperparameters_particle_dimer}). The trained Boltzmann
Generator has learned a transformation of the complex configuration
space density to a concentrated, 76-dimensional ball in latent space
(Fig. \ref{fig:particle_dimer}c). Indeed, direct sampling from a
76-dimensional Gaussian in latent space and transformation via $F_{zx}$
generates configurations where all particles are placed without significant
clashes, and potential energies that overlap with the energy distribution
of the unbiased MD trajectories (Fig. \ref{fig:particle_dimer}d).
Also, realistic transition states that have not been included in any
training data are sampled (Fig. \ref{fig:particle_dimer}d, middle).

To demonstrate the computation of thermodynamic quantities, we perform
training by energy (\ref{eq:main_loss_KL}) simultaneously to a range
of temperatures (Supp. Mat.). While the temperature changes the configuration
space distribution in a complex way it can be modeled as a simple
scaling factor in the width of the Gaussian prior distribution (Methods).
Then, we sample the Boltzmann Generator for a range of temperatures
and use simple reweighting to compute the free energies along the
dimer distances. As shown in Fig. \ref{fig:particle_dimer}e, these
temperature-dependent free energies agree precisely with extensive
umbrella sampling simulations that employ bias potentials along the
dimer distance (\cite{Torrie_JCompPhys23_187}, Supp. Mat.).

We estimate that the MD simulation needs at least $10^{12}$ steps
to spontaneously sample a single transition from closed to open state
and back (Supp. Mat.), and about $100$ such transitions would be
needed to compute free energy differences with the precision of Boltzmann
Generators shown in Fig. \ref{fig:particle_dimer}e. The total effort
to train the Boltzmann Generator is about $3\cdot10^{7}$ energy evaluations,
but then statistically independent samples can be drawn in one shot
at the entire temperature range trained at, resulting in at least
$7$ orders of magnitude speedup compared to MD.

As above, we perform linear interpolations between the latent space
representations of open- and closed-dimer samples. A significant fraction
of all pair interpolations result in low-energy pathways. The lowest-energy
interpolation of 100 randomly selected pairs of end-states is shown
in Fig. \ref{fig:particle_dimer}f, representing a physically meaningful
rearrangement of the dimer and solvent.
\begin{center}
\begin{figure}[H]
\begin{centering}
\includegraphics[width=0.65\textwidth]{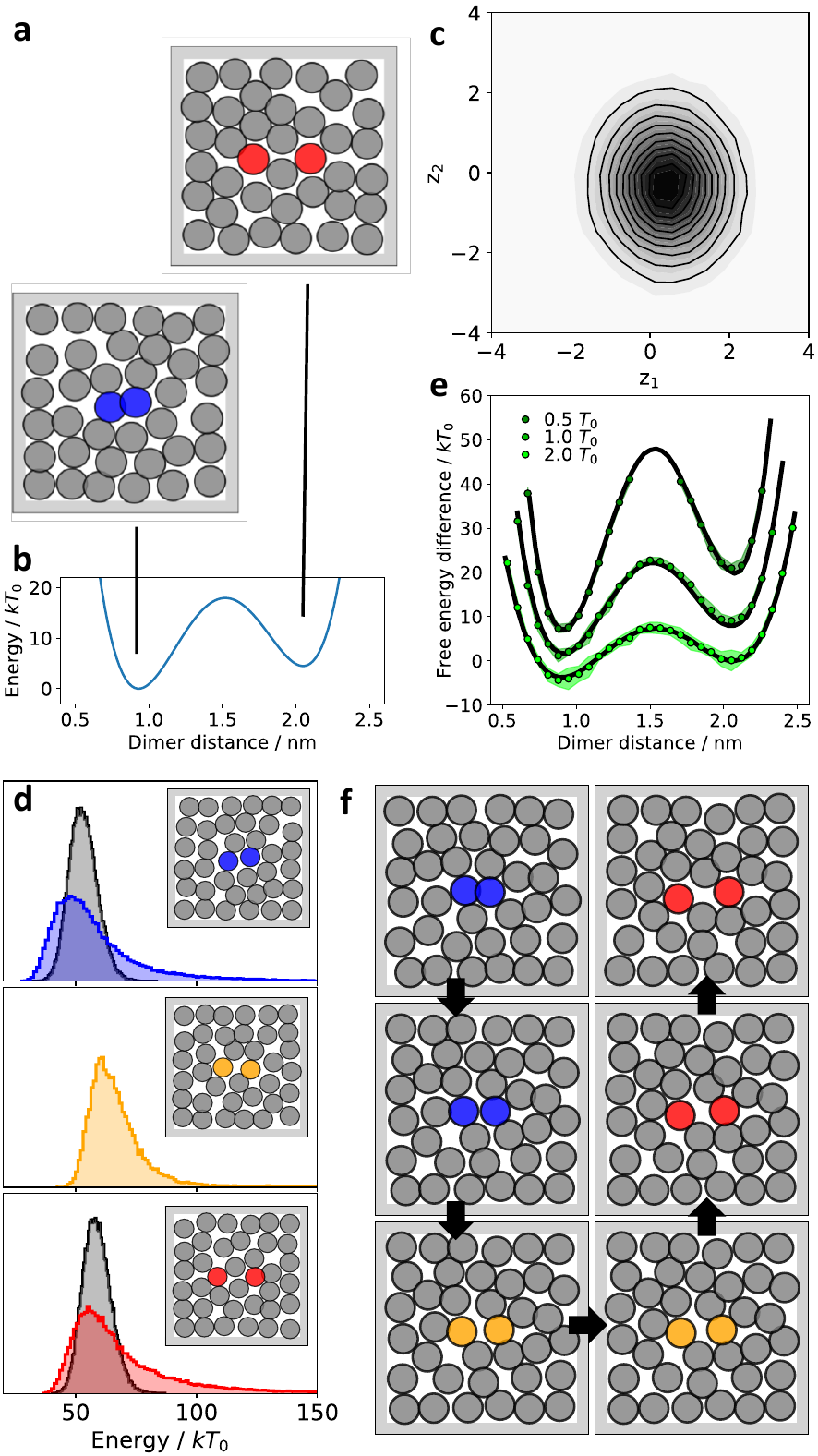}
\par\end{centering}
\caption{\label{fig:particle_dimer}\textbf{Repulsive particle System with
bistable dimer.} \textbf{a)} Closed (blue) and open (red) configurations
from MD simulations (input data). \textbf{b)} Bistable dimer potential.
\textbf{c)} Distribution of MD simulation data in latent space coordinates
$z_{1},z_{2}$ after training Boltzmann Generator. \textbf{d)} Potential
energy distribution from MD (grey) and Boltzmann generator for closed
(blue), open (red) and transition configurations (yellow). Insets
show one-shot Boltzmann Generator samples. \textbf{e)} Free energy
differences as a function of dimer distance and relative temperature
sampled with Boltzmann generators (generation and reweighting, green
bullets with intervals indicating one standard error from 10 independent
repeats) and umbrella sampling (black lines). \textbf{f}) Linear latent
space interpolation between the the closed and open structures shown
in top row.}
\end{figure}
\par\end{center}

\subsubsection*{Exploring configuration space}

In the previous examples, Boltzmann Generators were used to sample
known regions of configuration space and compute statistics thereof.
Here, we demonstrate that Boltzmann Generators can help to explore
configuration space. The basic idea is as follows: we construct an
exploratory sampling method by employing an established sampling algorithm
in latent space, while simultaneously training the Boltzmann Generator
transformation using the configurations found so far.

We initialize the method with a (possibly small) set of configurations
$X$. Training is done here by minimizing the symmetric loss function
$J=J_{KL}+J_{ML}$ (Eq. \ref{eq:main_loss_KL}-\ref{eq:main_loss_ML}).
The likelihood loss function $J_{ML}$ is initially biased by the
input data, but as $X$ approaches an unbiased Boltzmann sample, the
symmetric loss converges to a meaningful distance of probabilities
(Methods). As an example, we here use Metropolis Monte Carlo in the
latent space Boltzmann Generator to update $X$ (Methods). The step-size
is chosen adaptively but reaches the order of the latent space distribution
width. Thus, large-scale configuration transitions in physical space
can be overcome in a single Monte Carlo step.

We now revisit the three previous examples and initialize $X$ with
only a single input configuration from the most stable state (Fig.
\ref{fig:exploration}a, d, g). The exploration method quickly fills
the local metastable states, and finds new metastable states within
a few $10^{5}$ energy calls, i.e., orders of magnitude faster than
direct MD (Fig. \ref{fig:exploration}b, e, h). This demonstrates
that Boltzmann Generators sample new, previously unseen states with
a significant probability, and that this ability can be turned into
exploring configuration space when past samples are stored and reused
for training.

The Metropolis Monte Carlo method causes the sample to converge towards
the Boltzmann distribution. However, we do not need to wait for this
method to be converged, as with sufficient samples in the states of
interest, the equilibrium free energies can be computed by employing
reweighting as in Figs. \ref{fig:model_systems}-\ref{fig:particle_dimer}
above.  While new states are found, the data-based loss $J_{ML}$
may increase and decrease again, while the Boltzmann Generator transformation
is updated to include these new states (Fig. \ref{fig:exploration}c,
f, i; top row). During training, the energy-based loss $J_{KL}$ decreases
steadily until the full Boltzmann distribution is sampled (Fig. \ref{fig:exploration}c,
f, i; middle row). We also observe that the Metropolis Monte Carlo
efficiency, defined by the product of step-length and acceptance rate,
tends to increase over time, although it may decrease temporarily
when more states are found (Fig. \ref{fig:exploration}c, f, i; top
row).

Due to the invertible transformation between latent and configuration
space, any sampling method that involves reweighting or Monte Carlo
acceptance steps can be reformulated in Boltzmann Generator latent
space, and potentially yield enhanced performance.
\begin{center}
\begin{figure}
\begin{centering}
\includegraphics[width=0.6\textwidth]{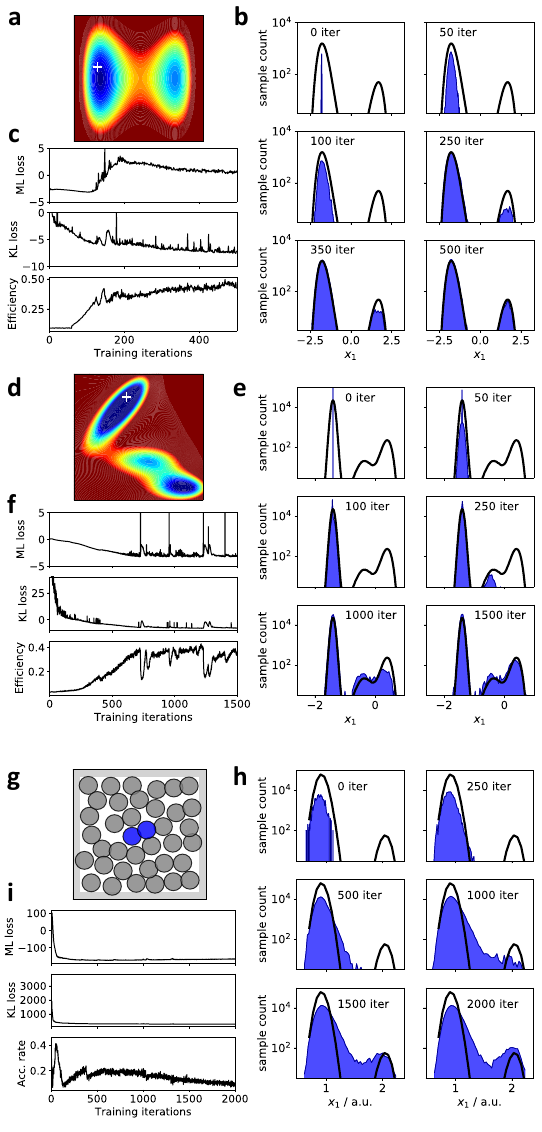}
\par\end{centering}
\vspace{-0.5cm}

\caption{\label{fig:exploration}\textbf{Exploration with Boltzmann Generators
from a single snapshot. a-c)} Double-well potential, \textbf{d}-\textbf{f})
Mueller potential, \textbf{g}-\textbf{h}) solvated particle dimer.
\textbf{a}, \textbf{d}, \textbf{g}) Starting configuration. \textbf{c},
\textbf{f}, \textbf{i}) Convergence of the loss terms ($J_{ML}$ and
$J_{KL}$) and the MCMC efficiency (product of step length and acceptance
rate). \textbf{b}, \textbf{e}, \textbf{h}) Evolution of sample distribution
over MCMC iteration. As soon as sufficient density is available in
the states of interest, these distributions can be reweighted to equilibrium
as in Figs. \ref{fig:model_systems}-\ref{fig:particle_dimer}.}
\end{figure}
\par\end{center}

\subsubsection*{Complex molecules}

We demonstrate that Boltzmann Generators can generate equilibrium
all-atom structures of macromolecules in one shot using the Bovine
Pancreatic Trypsin Inhibitor protein (BPTI) in an implicit solvent
model (Fig. \ref{fig:bpti}, Supp. Mat.). In order to train Boltzmann
Generators for complex molecular models, we wrapped the energy and
force computation functions of the OpenMM simulation software \cite{EastmanEtAl_JCTC13_OpenMM}
in the standard deep learning library Tensorflow \cite{tensorflow2015}.

Training a Boltzmann Generator directly on the Cartesian coordinates
resulted in large energies, and unrealistic structures with distorted
bond lengths and angles. This problem was solved by incorporating
the following coordinate transformation in the first layer of the
Boltzmann Generator that is invertible up to rotation and translation
of the molecule (Fig. \ref{fig:bpti}a, Methods): the coordinates
are split into Cartesian and internal coordinate sets. The Cartesian
coordinates include heavy atoms of the backbone and the disulfide
bridges. Cartesian coordinates are whitened, i.e. decorrelated and
normalized, using a principal component analysis of the input data.
During whitening, the six degrees of freedom corresponding to global
translation and rotation of the molecule are discarded. The remaining
side-chain atoms are measured in internal coordinates (bond-lengths,
angles and torsions with respect to parent atoms), and subsequently
normalized.

After this coordinate transformation, the learning problem is substantially
simplified as the transformed input data are already nearly Gaussian
distributed. We first demonstrate that Boltzmann Generators can learn
to sample heterogeneous equilibrium structures when trained with examples
from different configurations. To this end, we generated six short
simulations of 20 nanoseconds each, starting from snapshots of the
well-known 1-millisecond simulation of BPTI produced on the Anton
supercomputer \cite{Shaw_Science10_Anton}. In the more common situation
that no ultra-long MD trajectory is available, the Boltzmann Generator
can be seeded with different crystallographic structures or homology
models. In order to promote simultaneous sampling of high- and low-probability
configurations, we included an RC loss using two slow collective coordinates
that have been used earlier in the analysis of this BPTI simulation
(Supp. Mat.), \cite{PerezEtAl_JCP13_TICA,SchererEtAl_JCTC15_EMMA2}.
Boltzmann Generators without use of reaction coordinates are discussed
in the subsequent section.

Indeed, a trained Boltzmann Generator with 8 invertible blocks can
sample all 892 atom positions (2676 dimensions) in one shot and produce
locally and globally valid structures (Fig. \ref{fig:bpti}c). The
potential energies of samples exhibit significant overlap with the
potential energy distribution of MD simulations (Fig. \ref{fig:bpti}d),
thus samples can be reweighted for free energy calculations. The probability
distributions of most bond lengths and valence angles is almost indistinguishable
from the distributions of the equilibrium MD simulations (Fig. \ref{fig:bpti}e).
The only exception is that the distributions of valence angles involving
sulfur atoms is slightly narrower.

The trained Boltzmann Generator learns to encode and sample significantly
different structures (Fig. \ref{fig:bpti}f). In particular, it generates
independent one-shot samples of the near-crystallographic structure
``X'' (1.4 $\text{�}$ mean backbone RMSD to crystal structure),
and the open ``O'' structure which involves significant changes
in flexible loops and repacking of side-chains (Fig. \ref{fig:bpti}g,h).
The X$\rightarrow$O transition has been sampled only once in the
millisecond Anton trajectory, which is consistent with the observation
of a millisecond-timescale ``major-minor'' state transition observed
in nuclear magnetic resonance (NMR) spectroscopy \cite{Grey2003}.
We note that X$\leftrightarrow$O transition states are not included
in the Boltzmann Generator training data.

Sampling such a transition and collecting statistics for it is challenging
for any existing simulation method. Brute force MD simulations would
require several milliseconds of simulation data. To employ enhanced
sampling, an order parameter or reaction coordinate able to drive
the complex rearrangement shown in Fig. \ref{fig:bpti}g-h would need
to be found, but since BPTI has multiple states with lifetimes on
the order of 10-100 microseconds \cite{Shaw_Science10_Anton,SchererEtAl_JCTC15_EMMA2},
the simulation time required for convergence would still be extensive.
The computation of free energy differences using Boltzmann Generators
will be discussed in the next section.
\begin{center}
\begin{figure}[H]
\begin{centering}
\includegraphics[width=1\textwidth]{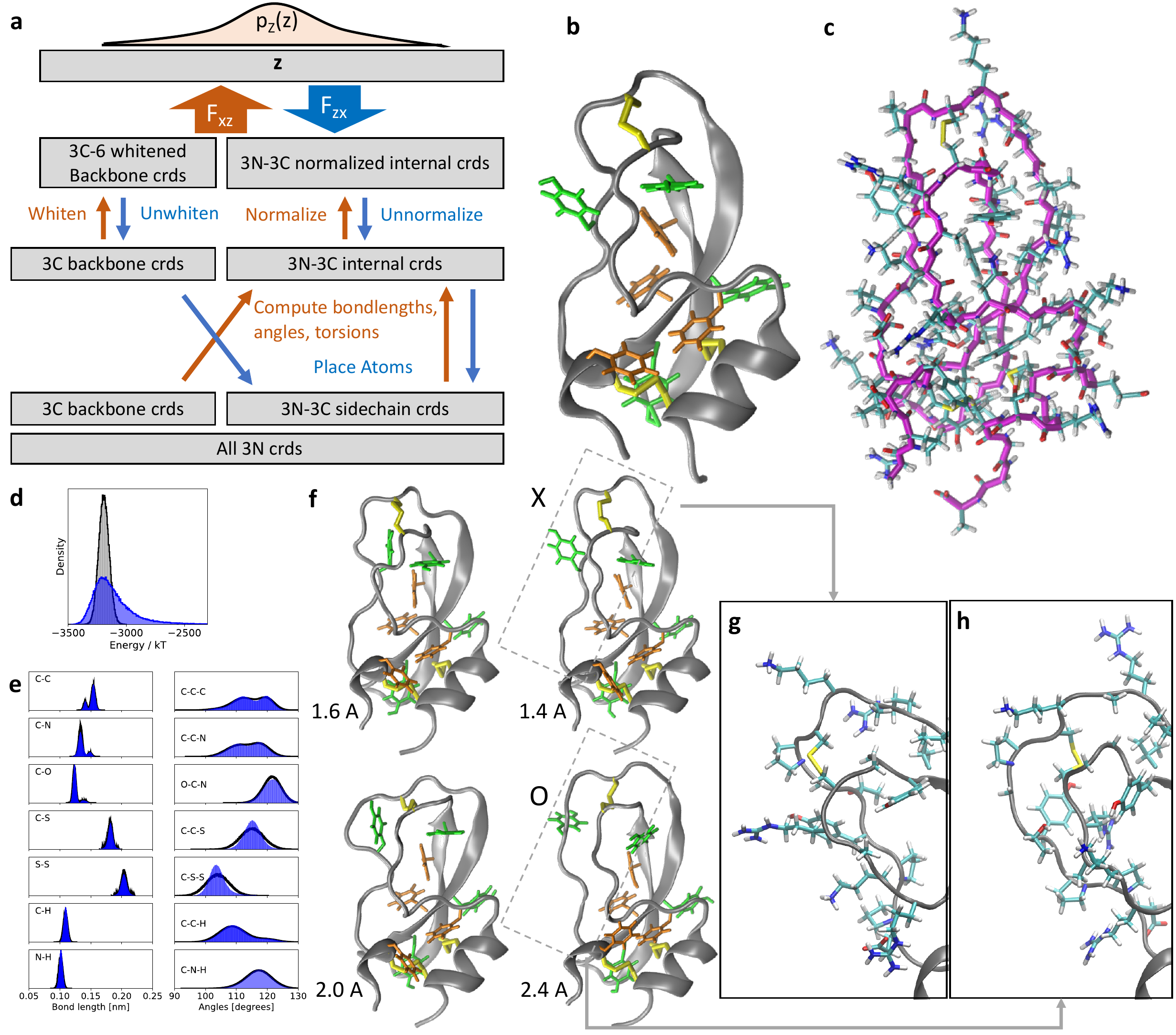}
\par\end{centering}
\caption{\label{fig:bpti}\textbf{One-shot sampling of all-atom structures
in different conformations of the BPTI protein}. \textbf{a}) Boltzmann
Generator for macromolecules: Backbone atoms are whitened using principal
component analysis, side chain atoms are described in normalized internal
coordinates. \textbf{b}) BPTI X-ray crystal structure (PDB: 5PTI).
Cysteine disulfide bridges and aromatic residues are shown for orientation.
\textbf{c}) One-shot Boltzmann Generator sample of all 892 atoms (2670
dimensions) of the BPTI protein similar to the X-ray structure. \textbf{d})
Potential energy distribution from MD simulation (grey) and Boltzmann
generator one-shot samples (blue). \textbf{e}) Distribution of bonds
and angles compared between MD simulation (black) and Boltzmann Generator
(blue). \textbf{f}) Representative snapshots of four clusters of structures
generated with the Boltzmann Generator. Backbone root mean square
deviation RMSDs from the X-ray structure is given below structure
(in $\text{�}$ngstr�m). Marked are the X-ray like structure ``X''
and the open structure ``O''. \textbf{g},\textbf{h}) Zoom into the
most variable parts of the Boltzmann-generated samples from the ``X''
and ``O'' states. Side-chains are shown in atomistic resolution.}
\end{figure}
\par\end{center}

\subsubsection*{Thermodynamics between disconnected states}

We develop a reaction-coordinate free approach to compute free energy
differences from disconnected MD or MCMC simulations in separated
states, such as two conformations of a protein. As demonstrated above,
this can be achieved by a single Boltzmann Generator that simultaneously
captures multiple metastable states and maps them to the same latent
space $Z$, where they are connected via the Gaussian prior distribution
(Fig. \ref{fig:model_systems}b,h, Fig. \ref{fig:particle_dimer}c).
However, a more direct statistical mechanics idea that has been successfully
applied to certain simple liquids and solids is to compute free energy
differences by relating to a tractable reference state, e.g., ideal
gas or crystal \cite{FrenkelLadd_JCP84_FreeEnergySolids,HooverRee_JCP68_MeltingTransition,YtrebergaZuckerman_JCP06_FreeEnergies}.
Here we show that Boltzmann Generators can turn this idea into a general
method applicable to complex many-body systems.

\begin{figure}[H]
\begin{centering}
\includegraphics[width=0.8\textwidth]{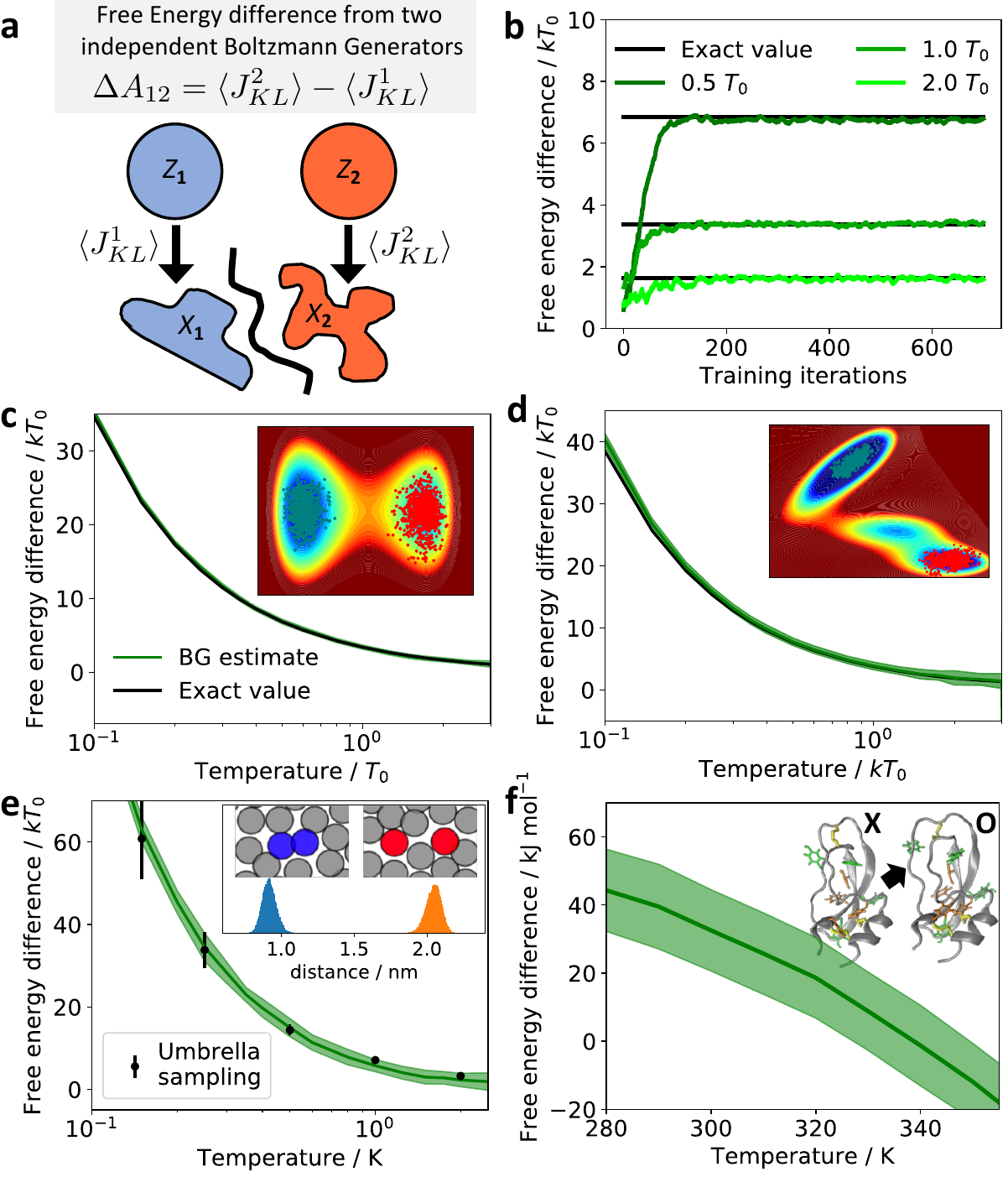}
\par\end{centering}
\caption{\label{fig:2BGfree_energy_diff}\textbf{Thermodynamics between disconnected
states by coupling multiple Boltzmann Generators}.\textbf{ a)} Using
multiple Boltzmann Generators, we can compute free energy differences
between states without requiring reaction coordinates, using only
disconnected MD simulations in each of them. This is possible because
each Boltzmann Generator estimates the free energy difference to a
common reference state. \textbf{b}) Example for tracking convergence:
estimate of free energy difference for the double well potential (multiple
temperatures) as a function of the training iterations of two Boltzmann
Generators. Convergence plots for the other systems are shown in Fig.
\ref{fig_convergence_2BG}. Results show estimates from two Boltzmann
Generators with mean and one standard error computed from bootstrapping
the converged segment of the free energy estimate. \textbf{c) }left-to-right
transition in the double well, \textbf{d)} left-to-right transition
in the Mueller potential, \textbf{e)} closed-to-open transition in
the solvated bistable particle dimer, \textbf{f)} X$\rightarrow$O
transition in an atomistic model of Bovine Pancreatic Trypsin Inhibitor
(BPTI).}
\end{figure}

Recall that the value of the energy loss function $J_{KL}$ (Eq. \ref{eq:main_loss_KL})
estimates the free energy difference of transforming the Gaussian
prior distribution to the generated distribution in configuration
space. If we are now given MD data sampled in two or more disconnected
states, we can train independent Boltzmann Generators for each of
them. The goal here is not to explore configuration space, so training
by energy is combined with training by example (Eq. \ref{eq:main_loss_ML})
in order to restrain the generated distribution around the separate
states. For each Boltzmann Generator, the transformation free energy
is computed, e.g., $\langle J_{KL}^{1}\rangle$ and $\langle J_{KL}^{2}\rangle$,
by sampling from the Gaussian prior distributions and inserting into
(\ref{eq:main_loss_KL}). The free energy difference between the two
states is directly given as a difference between these two values,
$\Delta A_{12}=\langle J_{KL}^{2}\rangle-\langle J_{KL}^{1}\rangle$
(Fig. \ref{fig:2BGfree_energy_diff}a).

We illustrate our method by computing temperature-dependent free energy
differences for the four systems discussed above, each using two completely
disconnected MD simulations as input. Since the estimate of the free
energy difference is readily available from the value of the loss
function, it can be conveniently tracked for convergence while the
Boltzmann Generators are trained (Fig. \ref{fig:2BGfree_energy_diff}b,
Fig. \ref{fig_convergence_2BG}).

For the two-dimensional systems (double well, Mueller potential) exact
reference values for the free energy differences can be computed,
and the Boltzmann Generator method recovers them accurately with a
small statistical uncertainty over the entire temperature range (Fig.
\ref{fig:2BGfree_energy_diff}c, d), using tenfold less simulation
data and about tenfold shorter training time than for the estimates
using a single joint Boltzmann Generator reported in Fig. \ref{fig:model_systems}
(Supp. Mat.).

For the solvated bistable dimer, we use simulations that are tenfold
shorter than for the single joint Boltzmann Generator reported in
Fig. \ref{fig:particle_dimer} and train two independent Boltzmann
Generators at multiple temperatures. As a reference, three independent
umbrella sampling simulations were conducted at each of five different
temperatures. Both predictions of the free energy difference between
open and closed dimer states are consistent and have overall similar
uncertainties (Fig. \ref{fig:2BGfree_energy_diff}e, Fig. \ref{fig_convergence_2BG}b,
note that the uncertainty of umbrella sampling is strongly temperature
dependent). Although Umbrella Sampling is well-suited for this system
with a clear reaction coordinate, the two-Boltzmann-Generator method
required 50 times less energy calls than the Umbrella Sampling simulations
at five temperatures, and yet makes predictions across the full temperature
range (Supp. Mat.).

Finally, we use the same method to predict the temperature-dependent
free energy difference of the ``X'' and ``O'' states in the BPTI
protein. Sampling this millisecond transition ten times by brute-force
MD would take around 30 years on one of the GTX1080 graphics cards
that are used for computations in this paper, and would only give
us the free-energy difference at a single temperature. While speeding
up this complex transition with enhanced sampling may be possible,
engineering a suitable order parameter is a time-consuming trial and
error task. Training two Boltzmann Generators does not require any
notion of reaction coordinate.

Here we use the two-Boltzmann-Generator method starting from two simulations
of 20 ns each, that were here started from selected frames of the
one-millisecond trajectory, but could generally be started from crystallographic
structures or homology models. Conducting the MD simulations, training
and analyzing the Boltzmann Generators used a total of less than $3\times10^{7}$
energy calls, which results in a converged free energy estimate within
a total of about 10 GPU hours, i.e. about 5 orders of magnitude faster
than the brute-force approach (Fig. \ref{fig:2BGfree_energy_diff}f,
Fig. \ref{fig_convergence_2BG}c). While no reference for this free
energy difference in the given simulation model is known, the temperature
profile admits basic consistency checks: The X-ray structure is identified
as the most stable structure at temperatures below 330 K. The internal
energy and entropy terms of the free energy difference (Eq. \ref{eq:main_loss_KL}),
are both positive across all temperatures. Consequently, the free
energy decreases at high temperatures as the entropic stabilization
becomes stronger. A higher configurational entropy of the ``O''
state is consistent with its more open loop structure (compare Fig.
\ref{fig:bpti}g and h) and the higher degree of fluctuations in the
``O'' state observed by the analysis in \cite{SchererEtAl_JCTC15_EMMA2}.

\section*{Discussion}

Boltzmann Generators can overcome rare event sampling problems in
many-body systems by generating independent samples from different
metastable states in one shot. We have demonstrated this for dense
and unstructured multi-body systems with up to 892 atoms (over 2600
dimensions) that are placed simultaneously, with most samples having
globally and locally valid structures and potential energies in the
range of the equilibrium distribution. In contrast to other generative
neural networks, Boltzmann Generators produce unbiased samples, as
the generated probability density is known at each sample point and
can be reweighted to the target Boltzmann distribution. This feature
directly translates into being able to compute free energy differences
between different metastable states.

In contrast to enhanced sampling methods that directly operate in
configuration space, such as Umbrella Sampling or Metadynamics, Boltzmann
Generators can sample between metastable states without any pre-defined
``reaction coordinate'' connecting them. This is achieved by learning
a coordinate transformation in which different metastable states become
neighbors in the transformed space where the sampling occurs. If suitable
reaction coordinates are known, these can be incorporated into the
training in order to sample continuous pathways between states, e.g.,
to compute a continuous free energy profile along a reaction.

As in many areas of machine learning, key to success is to choose
a representation of the input data which supports the learning problem.
For macromolecules, we have found that a successful representation
is to describe its backbone in Cartesian coordinates and all atoms
that branch off from the backbone in internal coordinates. Additionally,
normalizing these coordinates to mean zero and variance one already
makes their probability distribution close to a Gaussian normal distribution,
thus simplifying the learning problem considerably.

We have shown, in principle, how explicit solvent systems, can be
treated. For this it is essential to build the physical invariances
into the learning problem. Specifically, we need to account for permutation
invariance: when two equivalent solvent molecules exchange positions,
the potential energy of the system is unchanged, and so is the Boltzmann
probability.

We have demonstrated scaling of Boltzmann Generators to 1000's of
dimensions. Generative networks in other fields have been able to
generate photorealistic images with $10^{6}$ dimensions in one shot
\cite{KarrasEtAl_ProgressiveGrowingGANs}. However, for the present
application the statistical efficiency, i.e. the usefulness of these
samples to compute equilibrium free energies, will decline with increasing
dimension. Scaling to systems with 100,000's of dimensions or more,
such as solvated atomistic models of large proteins, can be achieved
in different ways. Sampling of the full atomistic system may be approached
by divide-and-conquer: In each iteration of such an approach, one
would re-sample the positions of a cluster of atoms using the sum
of potential energies between cluster atoms and all system atoms,
and then, e.g., perform Monte Carlo steps using these cluster proposals.
Alternatively, Boltzmann Generators could be used to sample lower-dimensional
free energy surfaces learnt from all-atom models \cite{ChenEtAl_PNAS15_FreeEnergyLearning}.

A caveat of Boltzmann Generators is that, depending on the training
method, they may not be ergodic, i.e. they may not be able to reach
all configurations. Here we have proposed a training methods that
promotes state space exploration by performing Monte Carlo steps in
the Boltzmann Generator's latent space while training the network.
This may be viewed as a general recipe: The whole plethora of existing
sampling algorithms, such as Umbrella Sampling, Metadynamics and replica
exchange, can be reformulated in the latent space of a Boltzmann Generator,
potentially leading to dramatic performance gains. Any such approach
can always be combined with MD or MCMC moves in configuration space
to ensure ergodicity.

Finally, the Boltzmann Generators described here learn a system-specific
coordinate transformation. The approach would become much more general
and efficient, if Boltzmann Generators could be pre-trained on certain
building blocks of a molecular system, such as oligopeptides in solvent
or a protein, and then re-used on a complex system consisting of these
building blocks. A promising approach is to involve transferrable
featurization methods developed in the context of machine learning
for quantum mechanics \cite{BehlerParrinello_PRL07_NeuralNetwork,RuppEtAl_PRL12_QML}.

In summary, Boltzmann Generators represent a powerful approach to
address the long-standing rare-event sampling problem in many-body
systems, and open the door for new developments in statistical mechanics.

\clearpage

\section*{Methods}

\subsection*{A. Invertible networks}

We employ invertible networks with trainable parameters $\boldsymbol{\theta}$
in order to learn the transformation between the Gaussian random variables
$\mathbf{z}$ and the Boltzmann-distributed random variables $\mathbf{x}$:
\begin{align*}
\mathbf{z} & =F_{xz}(\mathbf{x};\boldsymbol{\theta})\\
\mathbf{x} & =F_{zx}(\mathbf{z};\boldsymbol{\theta}).
\end{align*}
Hence $F_{xz}=F_{zx}^{-1}$. Each transformation has a Jacobian matrix
with the pairwise first derivatives of outputs with respect to inputs:
\begin{align*}
\mathbf{J}_{zx}(\mathbf{z};\boldsymbol{\theta}) & =\left[\frac{\partial F_{zx}(\mathbf{z};\boldsymbol{\theta})}{\partial z_{1}},...,\frac{\partial F_{zx}(\mathbf{z};\boldsymbol{\theta})}{\partial z_{n}}\right]\\
\mathbf{J}_{xz}(\mathbf{x};\boldsymbol{\theta}) & =\left[\frac{dF_{xz}(\mathbf{x};\boldsymbol{\theta})}{dx_{1}},...,\frac{dF_{xz}(\mathbf{x};\boldsymbol{\theta})}{dx_{n}}\right]
\end{align*}
The absolute value of the Jacobian's determinant, $\left|\det\mathbf{J}_{zx}(\mathbf{z};\boldsymbol{\theta})\right|$,
measures how much a volume element at $\mathbf{z}$ is scaled by the
transformation. Below we will omit the symbol $\boldsymbol{\theta}$
and use the abbreviations:
\begin{align*}
R_{xz}(\mathbf{x}) & =\left|\det\mathbf{J}_{xz}(\mathbf{x})\right|\\
R_{zx}(\mathbf{z}) & =\left|\det\mathbf{J}_{zx}(\mathbf{z})\right|.
\end{align*}

We use invertible transformations because they allow us to transform
random variables as follows:
\begin{align}
p_{X}(\mathbf{x}) & =p_{Z}(\mathbf{z})R_{zx}(\mathbf{z})^{-1}=p_{Z}(F_{xz}(\mathbf{x}))R_{xz}(\mathbf{x})\label{eq:transform_zx}\\
p_{Z}(\mathbf{z}) & =p_{X}(\mathbf{x})R_{xz}(\mathbf{x})^{-1}=p_{X}(F_{zx}(\mathbf{z}))R_{zx}(\mathbf{z})\label{eq:transform_xz}
\end{align}

\paragraph{Trainable invertible layers}

We employ the RealNVP transformation as trainable part of invertible
networks \cite{DinhBengio_RealNVP}. The main idea is to split the
variables into two channels, $\mathbf{x}=(\mathbf{x}_{1},\mathbf{x}_{2})$,
$\mathbf{z}=(\mathbf{z}_{1},\mathbf{z}_{2})$, and do only trivially
invertible operations on each channel, such as multiplication and
addition. Additionally, we use arbitrary, non-invertible artificial
neural networks $S$ and $T$ are that respectively scale and translate
the second input channel $\mathbf{x}_{2}$ using a nonlinear transformation
of the first input channel $\mathbf{x}_{1}$.
\begin{eqnarray}
f_{xz}(\mathbf{x}_{1},\mathbf{x}_{2}) & : & \begin{cases}
\mathbf{z}_{1}=\mathbf{x}_{1}\\
\mathbf{z}_{2}=\mathbf{x}_{2}\odot\exp\left(S(\mathbf{x}_{1};\boldsymbol{\theta})\right)+T(\mathbf{x}_{1};\boldsymbol{\theta})
\end{cases}\label{eq:RNVP_Fxz}\\
\log R_{xz} & = & \sum_{i}S_{i}(\mathbf{x}_{1};\boldsymbol{\theta})\label{eq:RNVP_Jxz}\\
f_{zx}(\mathbf{z}_{1},\mathbf{z}_{2}) & : & \begin{cases}
\mathbf{x}_{1}=\mathbf{z}_{1}\\
\mathbf{x}_{2}=\left(\mathbf{z}_{2}-T(\mathbf{x}_{1};\boldsymbol{\theta})\right)\:\:\:\odot\exp\left(-S(\mathbf{z}_{1};\boldsymbol{\theta})\right)
\end{cases}\label{eq:RNVP_Fzx}\\
\log R_{zx} & = & -\sum_{i}S_{i}(\mathbf{z}_{1};\boldsymbol{\theta})\label{eq:RNVP_Jzx}
\end{eqnarray}
A RealNVP ``block'' is defined two stacked RealNVP layers with channels
swapped, such that both channels are transformed:
\begin{align*}
(\mathbf{y}_{1},\mathbf{y}_{2}) & =f_{xy}(\mathbf{x}_{1},\mathbf{x}_{2})\\
(\mathbf{z}_{1},\mathbf{z}_{2}) & =f_{yz}(\mathbf{y}_{2},\mathbf{y}_{1})
\end{align*}
Boltzmann Generators are built by putting the forward and the inverse
of such blocks in parallel that share the same nonlinear transformations
$T$ and $S$ and parameters (Fig. \ref{fig:illustration}f).

\paragraph{PCA Whitening layer}

We define a fixed-parameter layer ``$W$'' in order to transform
the input coordinates into whitened principal coordinates. For systems
with roto-translationally invariant energy, we first remove global
translation and rotation by superimposing each configuration to a
reference configuration. We then perform principal component analysis
(PCA) on input coordinates $X$ by solving the eigenvalue problem
\[
X^{T}X\mathbf{R}=\mathbf{R}\boldsymbol{\Lambda}
\]
where $\mathbf{R}=[\mathbf{r}_{1},...,\mathbf{r}_{N}]$ are principal
components vectors and $\boldsymbol{\Lambda}=\mathrm{diag}(\lambda_{1},...,\lambda_{d})$
their variances. For systems with roto-translationally invariant energy,
the six smallest eigenvalues are 0 and are discarded along with the
corresponding eigenvectors. The whitening transformation and its inverse
are defined by:
\begin{align*}
W(\mathbf{x}): & \mathbf{z}=\boldsymbol{\Lambda}^{-\frac{1}{2}}\mathbf{R}^{\top}\mathbf{x}\\
W^{-1}(\mathbf{z}): & \mathbf{x}=\mathbf{R}\boldsymbol{\Lambda}^{\frac{1}{2}}\mathbf{z}
\end{align*}
Note that when translation and rotation are removed in the transformation,
this layer is only invertible for $\mathbf{x}$ where translation
and rotation are removed as well. However, the network is always invertible
for the relevant sequence $\mathbf{z}\rightarrow\mathbf{x}\rightarrow\mathbf{z}$.
The Jacobians of $W$ are:
\begin{align*}
\log R_{xz} & =-\frac{1}{2}\sum_{i}\log\lambda_{i}\\
\log R_{zx} & =\frac{1}{2}\sum_{i}\log\lambda_{i}
\end{align*}

\paragraph{Mixed Coordinate transformation layer}

In order to treat macromolecules we defined a new transformation layer
``$M$'' that transforms into mixed whitened Cartesian / normalized
internal coordinates. We first split the coordinates into a Cartesian
and an internal coordinate set, $\mathbf{x}\rightarrow[\mathbf{x}_{C},\mathbf{x}_{I}]$.
$\mathbf{x}_{C}$ is whitened (see above), $\mathbf{x}_{I}$, is transformed
into internal coordinates (ICs). For every particle $i$ in $\mathbf{x}_{I}$
we define three ``parent'' particles $j,k,l$, and the Cartesian
coordinates of particles $i,j,k,l$ are converted into distance, angle
and dihedral $(d_{ij},\alpha_{ijk},\phi_{ijkl})$. Finally, each IC
is normalized by subtracting the mean and dividing by the standard
deviation of the corresponding coordinates in the input data (Fig.
\ref{fig:bpti}a). PCA whitening and IC normalization are essential
for training Boltzmann Generators for complex molecules, as this sets
large fluctuations of the whole molecule on the same scale as small
vibrations of stiff coordinates such as bond lengths. We briefly call
the transformation to normalized internal coordinates $I(\mathbf{x})$.

The inverse transformation is straightforward: The Cartesian set is
first restored by applying $W^{-1}$. Then the particles in the internal
coordinate unnormalized and then placed in a valid sequence, i.e.
first particles $i$ whose parent particles are all in the Cartesian
set, then particles whose parents have just been placed, etc. As the
$W$ layer, the $M$ layer is invertible up to global translation
and rotation of the molecule that may have been removed during whitening.
Additionally, we prevent non-invertibility in dihedral space by avoiding
to generate angle values outside the range $[-\pi,\pi]$ (Suppl. Mat.).

The Jacobians of the $M$ layer are computed using Tensorflow's automatic
differentiation methods.

\subsection*{B. Training and using Boltzmann Generators}

The Boltzmann Generator is trained by minimizing a loss functional
of the following form:
\begin{equation}
J=w_{ML}J_{ML}+w_{KL}J_{KL}+w_{RC}J_{RC}.\label{eq:total_loss_weighted}
\end{equation}
where the terms represent maximum-likelihood (ML, ``training by example''),
Kullback-Leiber (KL, ``training by energy''), and reaction-coordinate
(RC) optimization and the $w$'s control their weights. Below we will
derive these terms in detail.

We call the ``exact'' distributions $\mu$ and the generated distributions
$q$. In particular, $\mu_{Z}(\mathbf{z})$ is the Gaussian prior
distribution from which we sample latent space variables and $q_{X}(\mathbf{x})$
is the distribution that results from the network transformation $F_{zx}$.
Likewise, $\mu_{X}(\mathbf{x})\propto\exp(-u(\mathbf{x}))$ is the
Boltzmann distribution in configuration space and $q_{Z}(\mathbf{z})$
is the distribution that results from the network transformation $F_{xz}$:
\begin{eqnarray*}
\mu_{Z}(\mathbf{z}) & \overset{F_{zx}}{\longrightarrow} & q_{X}(\mathbf{x})\\
\mu_{X}(\mathbf{x}) & \overset{F_{xz}}{\longrightarrow} & q_{Z}(\mathbf{z})
\end{eqnarray*}

\textbf{Boltzmann distribution}: A special case is to use Boltzmann
Generators to sample from the Boltzmann distribution of the canonical
ensemble. Other ensembles can be modeled by incorporating the choice
of ensemble into the reduced potential \cite{ShirtsChodera_JCP08_MBAR}.
The Boltzmann distribution has the form:
\begin{equation}
\mu_{X}(\mathbf{x})=Z_{X}^{-1}\mathrm{e}^{-\beta U(\mathbf{x})}\label{eq:Boltzmann_distribution}
\end{equation}
where $\beta^{-1}=k_{B}T$ with Boltzmann constant $k_{B}$ and temperature
$T$. When we only have one temperature, we define the reduced energy
\[
u(\mathbf{x})=\frac{U(\mathbf{x})}{k_{B}T}
\]
In order to work with multiple temperatures $(T^{1},...,T^{K})$,
we define a reference temperature $T^{0}$ and reduced energy $u^{0}(\mathbf{x})=U(\mathbf{x})/k_{B}T^{0}$.
The reduced energies are then obtained by scaling with the relative
temperature $\tau_{k}=T^{k}/T^{0}$:
\[
u^{k}(\mathbf{x})=\frac{T^{0}}{T^{k}}u^{0}(\mathbf{x})=\frac{u^{0}(\mathbf{x})}{\tau_{k}}.
\]

\textbf{Prior distribution}: We sample the input in $\mathbf{z}$
from the isotropic Gaussian distribution:
\begin{equation}
\mu_{Z}^{k}(\mathbf{z})=\mathcal{N}(\mathbf{0},\sigma_{k}^{2}\mathbf{I})=Z_{Z}^{-1}\mathrm{e}^{-\frac{1}{2}\left\Vert \mathbf{z}\right\Vert ^{2}/\sigma_{k}^{2}},\label{eq:z_Gaussian_prior}
\end{equation}
with normalization constant $Z_{Z}$. This corresponds to the prior
energy of a harmonic oscillator:
\begin{align}
u_{Z}^{k}(\mathbf{z}) & =-\log\mu_{Z}^{k}(\mathbf{z})\nonumber \\
 & =\frac{1}{2\sigma_{k}^{2}}\left\Vert \mathbf{z}\right\Vert ^{2}+\mathrm{const}.\label{eq:z_Gaussian_energy}
\end{align}
Thus the variance takes the same role as the relative temperature.
We (arbitrarily) choose variance 1 for the standard temperature, and
obtain: 
\[
\sigma_{k}^{2}=\tau_{k}.
\]

\paragraph{Latent KL divergence}

The KL divergence measures the difference between two distributions
$q$ and $p$:
\begin{align*}
\mathrm{KL}(q\parallel p) & =\int q(\mathbf{x})\left[\log q(\mathbf{x})-\log p(\mathbf{x})\right]\mathrm{d}\mathbf{x},\\
 & =-H_{q}-\int q(\mathbf{x})\log p(\mathbf{x})\mathrm{d}\mathbf{x},
\end{align*}
where $H_{q}$ is the entropy of distribution $q$. Here we minimize
the difference between the probability densities predicted by the
Boltzmann generator and the respective reference distribution. Using
Equations (\ref{eq:transform_zx},\ref{eq:transform_xz},\ref{eq:Boltzmann_distribution})
the KL divergence in latent space is:
\begin{align*}
\mathrm{KL}_{\boldsymbol{\theta}}\left[\mu_{Z}\parallel q_{Z}\right] & =-H_{Z}-\int\mu_{Z}(\mathbf{z})\log q_{Z}(\mathbf{z};\boldsymbol{\theta})\mathrm{d}\mathbf{z},\\
 & =-H_{Z}-\int\mu_{Z}(\mathbf{z})\left[\log\mu_{X}(F_{zx}(\mathbf{z};\boldsymbol{\theta}))+\log R_{zx}(\mathbf{z};\boldsymbol{\theta})\right]\mathrm{d}\mathbf{z},\\
 & =-H_{Z}+\log Z_{X}+\mathbb{E}_{\mathbf{z}\sim\mu_{Z}(\mathbf{z})}\left[u(F_{zx}(\mathbf{z};\boldsymbol{\theta}))-\log R_{zx}(\mathbf{z};\boldsymbol{\theta})\right]
\end{align*}
Here, $\boldsymbol{\theta}$ are the trainable neural network parameters.
Since $H_{Z}$ and $Z_{X}$ are constants in $\boldsymbol{\theta}$,
the KL loss is given by:
\begin{equation}
J_{KL}=\mathbb{E}_{\mathbf{z}\sim\mu_{Z}(\mathbf{z})}\left[u(F_{zx}(\mathbf{z};\boldsymbol{\theta}))-\log R_{zx}(\mathbf{z};\boldsymbol{\theta})\right].\label{eq:loss_KL}
\end{equation}
Practically, each training batch samples points $\mathbf{z}\sim q_{Z}(\mathbf{z})$
from a normal distribution, transforms them via $F_{zx}$, and evaluates
Eq. (\ref{eq:loss_KL}). As shown in the Supp. Mat., the KL loss can
be rewritten to:
\begin{equation}
J_{KL}=U-H_{X}+H_{Z}\label{eq:J_KL_free_energy}
\end{equation}
which is, up to the constant $H_{Z}$ equal to the free energy of
the generated distribution with enthalpy $U$ and entropic factor
$H_{X}$.

We can extend (\ref{eq:loss_KL}) to simultaneously train at multiple
temperatures, obtaining:
\[
J_{KL}^{T^{1},...,T^{K}}=\sum_{k=1}^{K}\mathbb{E}_{\mathbf{z}\sim\mu_{Z}^{k}(\mathbf{z})}\left[u^{k}(F_{zx}(\mathbf{z};\boldsymbol{\theta}))-\log R_{zx}(\mathbf{z};\boldsymbol{\theta})\right].
\]
The KL divergence $\mathrm{KL}_{\boldsymbol{\theta}}\left[\mu_{Z}\parallel q_{Z}\right]$
is also minimized in probability density distillation used in different
contexts, e.g. in the training of recent audio generation networks
\cite{VanDenOord_WaveNet2}.

\paragraph{Reweighting and interpretation of latent KL as reweighting loss}

A simple way to compute quantitative statistics using Boltzmann generators
is to employ reweighting of probability densities, by assigning the
statistical weight $w_{X}(\mathbf{x})$ to each generated configuration
$\mathbf{x}$. Using Eq. (\ref{eq:transform_zx}-\ref{eq:transform_xz}),
we obtain:
\begin{align}
w_{X}(\mathbf{x}) & =\frac{\mu_{X}(\mathbf{x})}{q_{X}(\mathbf{x})}=\frac{q_{Z}(\mathbf{z})}{\mu_{Z}(\mathbf{z})}.\label{eq:reweighting_w}\\
 & \propto\mathrm{e}^{-u_{X}\left(F_{zx}(\mathbf{z})\right)+u_{Z}(\mathbf{z})+\log R_{zx}(\mathbf{z};\boldsymbol{\theta})}\nonumber 
\end{align}
Equilibrium expectation values can then be computed as
\begin{equation}
\mathbb{E}[O]\approx\frac{\sum_{i=1}^{N}w_{X}(\mathbf{x})O(\mathbf{x})}{\sum_{i=1}^{N}w_{X}(\mathbf{x})}.\label{eq:weighted_expectation}
\end{equation}

All free energy profiles shown in Figs. \ref{fig:model_systems},
\ref{fig:particle_dimer} and Fig. \ref{fig_training_methods_RealNVP}
were computed by $-k_{B}T\log p(R(\mathbf{x}))$ where $p(R(\mathbf{x}))$
is a probability density computed from a weighted histogram of the
coordinate $R(\mathbf{x})$ using the weighted expectation (\ref{eq:weighted_expectation}).
Histogram bins with weights worth less than 0.01 samples are discarded
to avoid making unreliable predictions.

Using Eq. (\ref{eq:reweighting_w}), it can be shown that minimization
of the KL divergence (\ref{eq:loss_KL}) is equivalent to maximizing
the sample weights:
\begin{align*}
\min\mathrm{KL}_{\boldsymbol{\theta}}\left[\mu_{Z}\parallel q_{Z}\right] & =\min\mathbb{E}_{\mathbf{z}\sim\mu_{Z}(\mathbf{z})}\left[\log\mu_{Z}(\mathbf{z})-\log q_{Z}(\mathbf{z};\boldsymbol{\theta})\right]\\
 & =\max\mathbb{E}_{\mathbf{z}\sim\mu_{Z}(\mathbf{z})}\left[\log w_{X}(\mathbf{x}\mid\mathbf{z})\right].
\end{align*}

\paragraph{Configuration KL divergence and Maximum Likelihood}

Likewise, we can express the KL divergence in $\mathbf{x}$ space
where we compute the divergence between the probability of generated
samples with their Boltzmann weight. Using Eqs. (\ref{eq:transform_zx},\ref{eq:transform_xz},\ref{eq:z_Gaussian_prior}):
\begin{align*}
\mathrm{KL}_{\boldsymbol{\theta}}\left[\mu_{X}\parallel q_{X}\right] & =H_{X}-\int\mu_{X}(\mathbf{x})\log q_{X}(\mathbf{x};\boldsymbol{\theta})\mathrm{d}\mathbf{x}\\
 & =H_{X}-\int\mu_{X}(\mathbf{x})\left[\log\mu_{Z}(F_{xz}(\mathbf{x};\boldsymbol{\theta}))+\log R_{xz}(\mathbf{z};\boldsymbol{\theta})\right]\mathrm{d}\mathbf{x}.\\
 & =H_{X}+\log Z_{Z}+\mathbb{E}_{\mathbf{x}\sim\mu(\mathbf{x})}\left[\frac{1}{\sigma^{2}}\left\Vert F_{xz}(\mathbf{x};\boldsymbol{\theta})\right\Vert ^{2}-\log R_{xz}(\mathbf{x};\boldsymbol{\theta})\right].
\end{align*}

This loss is difficult to evaluate because we cannot sample from $\mu(\mathbf{x})$
a priori. However we can approximate the configuration KL divergence
by starting from a sample $\rho(\mathbf{x})$, resulting in:
\begin{align*}
J_{ML} & =-\mathbb{E}_{\mathbf{x}\sim\rho(\mathbf{x})}\left[\log q_{X}(\mathbf{x};\boldsymbol{\theta})\right]\\
 & =\mathbb{E}_{\mathbf{x}\sim\rho(\mathbf{x})}\left[\frac{1}{\sigma^{2}}\left\Vert F_{xz}(\mathbf{x};\boldsymbol{\theta})\right\Vert ^{2}-\log R_{xz}(\mathbf{x};\boldsymbol{\theta})\right]
\end{align*}
$J_{ML}$ is the negative log-likelihood, i.e. minimizing it maximizes
the likelihood of the sample $\rho(\mathbf{x})$ in the Gaussian prior
density.

\paragraph{Symmetric divergence}

The two KL divergences above can be naturally combined to the symmetric
divergence
\[
\mathrm{KL}_{\mathrm{sym}}=\frac{1}{2}\mathrm{KL}\left[\mu_{X}\parallel q_{X}\right]+\frac{1}{2}\mathrm{KL}\left[\mu_{Z}\parallel q_{Z}\right]
\]
which corresponds, up to an additive constant, to the Jensen-Shannon
divergence which uses the geometric mean of $m=\sqrt{q_{X}q_{Z}}$
instead of the arithmetic mean.

\paragraph{Reaction coordinate loss}

In some applications we do not want to sample from the Boltzmann distribution
but promote the sampling of high-energy states in a specific direction
of configuration space, for example in order to compute a free energy
profile along a predefined reaction coordinate $r(\mathbf{x})$ (Fig.
\ref{fig:model_systems}e,k). This is achieved by adding the reaction-coordinate
(RC) loss to the minimization problem:
\begin{align*}
J_{RC} & =\int p\left(r(\mathbf{x})\right)\log p\left(r(\mathbf{x})\right)\:\mathrm{d}r(\mathbf{x})\\
 & =\mathbb{E}_{\mathbf{x}\sim q_{X}(\mathbf{x})}\log p\left(r(\mathbf{x})\right).
\end{align*}
To implement this loss, the function $r$ is a user input, minimum
and maximum bounds are given, and $p\left(r(\mathbf{x})\right)$ is
computed as a batch-wise kernel density estimate along between the
bounds.

\subsection*{C. Adaptive sampling and training}

We define the following adaptive sampling method that trains a Boltzmann
Generator while simultaneously using it to propose new samples. The
method has a sample buffer $X$ that stores a pre-defined number of
$\mathbf{x}$ samples. This number is chosen such that low-probability
states of interest still have a chance to be part of the buffer when
it represents an equilibrium sample. For the examples in Fig. \ref{fig:exploration}
it was chosen to be 10,000 (double well, Mueller potential) and 100,000
(solvated particle dimer). $X$ can be initialized with any candidates
for configurations, in the examples in Fig. \ref{fig:exploration}
it was initialized with only one configuration (copied to all elements
of $X$), for the particle dimer system we additionally added small
Gaussian noise with standard deviation 0.05 nm to avoid that the initial
Boltzmann Generator overfits on a single point. The Boltzmann Generator
was initially trained by example, minimizing $J_{ML}$, using batch-size
128 and 20, 20, and 200 iterations for double well, Mueller potential,
and particle dimer, respectively. We then iterated the following adaptive
sampling and training loop using batch-size $1000$ for all examples.
\begin{enumerate}
\item Sample batch $\{\mathbf{x}_{1},...,\mathbf{x}_{B}\}$ from $X$.
\item Update Boltzmann Generator parameters $\boldsymbol{\theta}$ by training
on batch.
\item For each $\mathbf{x}$ in batch, propose a Metropolis Monte Carlo
step in latent space with step-size $s$:
\[
\mathbf{z}'=T_{xz}(\mathbf{x})+s\mathcal{N}(\mathbf{0},\mathbf{I}).
\]
\item Accept or reject proposal with probability $\min\{1,\exp(-\Delta E)\}$
using:
\[
\Delta E=u(T_{zx}(\mathbf{z}'))-u(\mathbf{x})-\log R_{zx}(\mathbf{z}';\boldsymbol{\theta})+\log R_{xz}(\mathbf{x};\boldsymbol{\theta})
\]
For the accepted samples, replace $\mathbf{x}$ by $\mathbf{x}'=T_{zx}(\mathbf{z}')$.
\end{enumerate}
\clearpage

\textbf{References and Notes:}

\paragraph{Acknowledgements}

We are grateful to Cecilia Clementi (Rice University), Brooke Husic,
Mohsen Sadeghi, Moritz Hoffmann (FU Berlin) and Phiala Shanahan (MIT)
for valuable comments and discussions.

\paragraph{Funding}

We acknowledge funding from European Commission (ERC CoG 772230 ``ScaleCell''),
Deutsche Forschungsgemeinschaft (CRC1114/A04, GRK2433 DAEDALUS), the
MATH$^{+}$ Berlin Mathematics research center (AA1x8, EF1x2) and
the Alexander von Humboldt foundation (Postdoctoral fellowship to
S.O.).

\paragraph{Author contributions}

F.N., S.O., J.K. designed and conducted research. F.N., J.K. and H.W.
developed theory. F.N., S.O., J.K. developed computer code. F.N. and
S.O. wrote the paper.

\paragraph{Competing interests}

The authors declare no competing interests.

\textbf{Data and materials availability}:

Data and computer code for generating results of this paper are available
at

http://doi.org/10.5281/zenodo.3242635

\clearpage

\setcounter{page}{1}
\begin{center}
{\huge{}Supplementary Materials for}{\huge\par}
\par\end{center}

\begin{center}
{\Large{}Boltzmann Generators -- Sampling Equilibrium States of Many-Body
Systems with Deep Learning}{\Large\par}
\par\end{center}

\begin{center}
Frank No\'{e}$^{1,2,3,\dagger,*}$, Simon Olsson$^{1,\dagger}$, Jonas
K\"ohler$^{1,\dagger}$ and Hao Wu$^{4,1}$
\par\end{center}

\begin{center}
correspondence to: frank.noe@fu-berlin.de
\par\end{center}

\textbf{Affiliations}:

$1$: FU Berlin, Department of Mathematics and Computer Science, Arnimallee
6, 14195 Berlin, Germany

$2$: FU Berlin, Department of Physics, Arnimallee 14, 14195 Berlin,
Germany

$3$: Rice University, Department of Chemistry, Houston, Texas 77005,
United States

$4$: Tongji University, School of Mathematical Sciences, Shanghai,
200092, P.R. China

$\dagger$: Equal contribution

\textbf{This PDF file includes}:
\begin{itemize}
\item Supplementary Text
\item Figures S1-3
\item Table S1
\item References 39-45
\end{itemize}
\clearpage

\subsection*{Supplementary Text}

\paragraph*{$M$ layer: ensuring invertibility in dihedral space}

For the mixed coordinate transformation layer $M$, it must be avoided
that the Boltzmann Generator samples internal coordinates that are
outside the range that are generated by the Cartesian coordinate transformation
$I(\mathbf{x})$ (by default $[-\pi,\pi]$ before normalization).
While angle values outside these bounds pose no problem for the placement
of atom positions as they are automatically periodically wrapped during
this process, they would break invertibility $\mathbf{z}\rightarrow\mathbf{x}\rightarrow\mathbf{z}$,
and thus invalidate the random variable transformation principle of
the Boltzmann Generator. Here we avoid this problem by adding a simple
quadratic loss during training that penalizes angles generated outside
the $[-\pi,\pi]$ range with a weight $w_{\mathrm{torsion}}$ and
is inactive within the range. This excludes violations of invertibility
for most, but not all samples. To ensure we are working on the manifold
which is invertible up to a global roto-translation we then simply
discard those samples for which invertibility $\mathbf{z}\rightarrow\mathbf{x}\rightarrow\mathbf{z}$
is violated.

\paragraph*{Derivation of the $KL$ loss as free energy}

For invertible transformation $F_{xz}$, we use the following relationship
of the entropies of the two distributions:
\begin{align}
H_{X} & =-\int_{\mathbf{x}}q_{X}(\mathbf{x})\log q_{X}(\mathbf{x})\:\mathrm{d}\mathbf{x}\nonumber \\
 & =-\int_{\mathbf{z}}q_{X}(F_{zx}(\mathbf{z}))\log\left(q_{X}(F_{zx}(\mathbf{z}))\,R_{zx}(\mathbf{z})\right)\:\mathrm{d}\mathbf{z}\nonumber \\
 & =-\int_{\mathbf{z}}\mu_{Z}(\mathbf{z})\log q_{X}(F_{zx}(\mathbf{z}))\:\mathrm{d}\mathbf{z}\nonumber \\
 & =-\int_{\mathbf{z}}\mu_{Z}(\mathbf{z})\log\left(\mu_{Z}(\mathbf{z})R_{zx}(\mathbf{z})^{-1}\right)\:\mathrm{d}\mathbf{z}\nonumber \\
 & =-\left(\int_{\mathbf{z}}\mu_{Z}(\mathbf{z})\log\mu_{Z}(\mathbf{z})\:\mathrm{d}\mathbf{z}\right)+\mathbb{E}_{\mathbf{z}\sim\mu_{Z}(\mathbf{z})}\log R_{zx}(\mathbf{z})\nonumber \\
 & =H_{Z}+\mathbb{E}_{\mathbf{z}\sim\mu_{Z}(\mathbf{z})}\left[\log R_{zx}(\mathbf{z})\right]\label{eq:entropy_difference}
\end{align}
Hence we have:
\begin{align*}
\mathrm{KL}_{\boldsymbol{\theta}}\left[\mu_{Z}\parallel q_{Z}\right] & =-H_{Z}+\log Z_{X}+\mathbb{E}_{\mathbf{z}\sim\mu_{Z}(\mathbf{z})}\left[u(F_{zx}(\mathbf{z};\boldsymbol{\theta}))\right]-\mathbb{E}_{\mathbf{z}\sim\mu_{Z}(\mathbf{z})}\left[\log R_{zx}(\mathbf{z};\boldsymbol{\theta})\right]\\
 & =-H_{X}+\log Z_{X}+\mathbb{E}_{\mathbf{z}\sim\mu_{Z}(\mathbf{z})}\left[u(F_{zx}(\mathbf{z};\boldsymbol{\theta}))\right]\\
 & =-H_{X}+\log Z_{X}+\mathbb{E}_{\mathbf{x}\sim\mu_{X}(\mathbf{x};\boldsymbol{\theta})}\left[u(\mathbf{x})\right]\\
 & =\mathrm{KL}_{\boldsymbol{\theta}}\left[q_{X}\parallel\mu_{X}\right].
\end{align*}
Then, the KL loss function becomes the free energy shown in Eq. (\ref{eq:J_KL_free_energy}).

\subsubsection*{Simulation systems and general hyper-parameter choices}

The ``MD'' simulations of the model systems (double well, Mueller
potential, solvated particle dimer) are not using actual molecular
dynamics, but are emulated with Metropolis Monte Carlo with small
local steps. In each step, a random vector from an isotropic Gaussian
distribution with a system-dependent standard deviation $\sigma_{\mathrm{Metro}}$
is added to the present configuration. This proposed configuration
is accepted or rejected with a standard Metropolis acceptance criterion.

All Boltzmann Generator networks are composed of invertible blocks
of non-volume preserving RealNVP layers. Each block contains two such
layers to make sure that all dimensions are subject to a nonlinear
transformation (Fig. \ref{fig:illustration}b). Each configuration
$\mathbf{x}$ or latent vector $\mathbf{z}$ is split into a channel
of ``even'' and ``odd'' dimensions, defining the pairs $(\mathbf{x}_{1},\mathbf{x}_{2})$
and $(\mathbf{z}_{1},\mathbf{z}_{2})$, respectively. To describe
the network architecture used, we use $R$ to denote RealNVP block
and $W$ for a PCA-based whitening layer. A subscript is used to denote
the number of repetitions of a motif, e.g. $R_{10}$ are ten stacked
RealNVP blocks.

We always used ReLU (rectified linear units) nonlinearities for the
translation networks ($T$ in Eq. \ref{eq:RNVP_Fxz},\ref{eq:RNVP_Fzx})
and tanh nonlinearities for the scaling networks ($S$ in Eq. \ref{eq:RNVP_Fxz},\ref{eq:RNVP_Fzx}).
For each Boltzmann Generator, $T$ and $S$ use equal network architectures
with $l_{\mathrm{hidden}}$ hidden layers containing $n_{\mathrm{hidden}}$
neurons each.

All networks are trained using the Adam adaptive stochastic gradient
descent method \cite{KingmaBa_ADAM}. Other choices and hyper-parameters
are described below.

In the first iterations of training that involves minimizing the loss
term $J_{KL}$, i.e. when the Boltzmann Generator first starts generating
samples whose free energies are being minimized, there is a significant
chance of generating extremely high energy values. We therefore regularize
the energy as follows:
\[
E_{\mathrm{reg}}=\begin{cases}
E & E<E_{\mathrm{high}}\\
E_{\mathrm{high}}+\log\left(E-E_{\mathrm{high}}+1\right) & E_{\mathrm{high}}\le E<E_{\mathrm{max}}\\
E_{\mathrm{high}}+\log\left(E_{\mathrm{max}}-E_{\mathrm{high}}+1\right) & E_{\mathrm{max}}<E
\end{cases}
\]
where $E_{\mathrm{max}}=10^{20}$ is a cutoff just set to avoid overflow
and $E_{\mathrm{high}}$ is initially very large and then gradually
reduced during training, but is left at a value far above the equilibrium
energies. The aim is that after training almost all samples end up
in the linear regime $E<E_{\mathrm{high}}$ where the employed loss
functions are meaningful.

\paragraph*{Double well}

We define a two-dimensional toy model which is bistable in $x$-direction
and harmonic in $y$-direction:
\begin{equation}
E(x,y)=\frac{1}{4}ax^{4}-\frac{1}{2}bx^{2}+cx+\frac{1}{2}dy^{2}\label{eq:double_well_energy}
\end{equation}
with $a=c=d=1$ and $b=6$ -- see Fig. \ref{fig:model_systems}a.
The system is simulated with a Metropolis step of $\sigma_{\mathrm{Metro}}=0.1$.
To estimate the average time needed for a return trip between both
states, we construct another systems with $a=0.25$ and $b=1.5$ that
has the same position of minima and the same energy difference between
them, but a much smaller barrier. For the ``flat'' systems frequent
transitions between the two end-states are observed. The return-trip
time of the original system is then estimated by $t=t_{\mathrm{flat}}\exp\left(B-B_{\mathrm{flat}}\right)$,
where $B,B_{\mathrm{flat}}$ are the energy barriers for the original
and the ``flat'' system from either one of the two minima, and $t,t_{\mathrm{flat}}$
are the times taken for a round-trip between the states. This results
in an estimate of $t=4\cdot10^{6}$ simulation steps for a return
trip in the double well system shown in Fig. \ref{fig:model_systems}a.

Boltzmann Generators for the double well system use the following
hyper-parameters and training schedules:
\begin{center}
\begin{tabular}{|c|c|c|c|c|c|}
\hline 
Results figure & Input samples & Network & $l_{\mathrm{hidden}}$ & $n_{\mathrm{hidden}}$ & temperatures\tabularnewline
\hline 
\hline 
Fig. \ref{fig:model_systems}a-f, Suppl. Fig. \ref{fig_training_methods_RealNVP} & 1000 & $R_{4}$ & 3 & 100 & 1.0\tabularnewline
\hline 
Fig. \ref{fig:2BGfree_energy_diff}c & 100 & $R_{4}$ & 3 & 100 & 0.5, 1.0, 2.0, 4.0\tabularnewline
\hline 
\end{tabular}
\par\end{center}

\begin{center}
\begin{tabular}{|c||c|c||c|c|}
\cline{2-5} \cline{3-5} \cline{4-5} \cline{5-5} 
\multicolumn{1}{c|}{} & \multicolumn{2}{c||}{Fig. \ref{fig:model_systems}a-f} & \multicolumn{2}{c|}{Fig. \ref{fig:2BGfree_energy_diff}c}\tabularnewline
\hline 
iter & 200 & 500 & 200 & 100\tabularnewline
\hline 
batch & 128 & 1000 & 128 & 1000\tabularnewline
\hline 
lr & 0.01 & 0.001 & 0.01 & 0.001\tabularnewline
\hline 
$w_{ML}$ & 1 & 1 & 1 & 1\tabularnewline
\hline 
$w_{KL}$ & 0 & 1 & 0 & 1\tabularnewline
\hline 
$w_{RC}$ & 0 & 0/1$^{*}$ & 0 & 0\tabularnewline
\hline 
\end{tabular}
\par\end{center}

$*$: $w_{RC}=0$ for ``green'' results and $w_{RC}=1$ for ``orange''
results in Fig. \ref{fig:model_systems}. For Fig. \ref{fig_training_methods_RealNVP}
different training schedules were compared, as described in the figure
caption.

\textbf{Mueller potential}

A scaled version of the Mueller potential was defined as:
\[
E(x,y)=\alpha\sum_{j=1}^{4}A_{j}\exp\left[a_{j}\left(x-\hat{x}_{j}\right)^{2}+b_{j}\left(x-\hat{x}_{j}\right)\left(y-\hat{y}_{j}\right)+c_{j}\left(y-\hat{y}_{j}\right)^{2}\right]
\]
with scaling parameter $\alpha=0.1$ and:
\begin{center}
\begin{tabular}{|c|c|c|c|c|}
\hline 
 & 1 & 2 & 3 & 4\tabularnewline
\hline 
\hline 
$a_{j}$ & -1 & -1 & -6.5 & 0.7\tabularnewline
\hline 
$b_{j}$ & 0 & 0 & 11 & 0.6\tabularnewline
\hline 
$c_{j}$ & -10 & -10 & 6.5 & 0.7\tabularnewline
\hline 
$A_{j}$ & -200 & -100 & -170 & 15\tabularnewline
\hline 
$\hat{x}_{j}$ & 1 & 0 & -0.5 & -1\tabularnewline
\hline 
$\hat{y}_{j}$ & 0 & 0.5 & 1.5 & 1\tabularnewline
\hline 
\end{tabular}
\par\end{center}

Boltzmann Generator architecture and training schedules were chosen
as follows:
\begin{center}
\begin{tabular}{|c|c|c|c|c|c|}
\hline 
Results figure & Input samples & Network & $l_{\mathrm{hidden}}$ & $n_{\mathrm{hidden}}$ & temperatures\tabularnewline
\hline 
\hline 
Fig. \ref{fig:model_systems}g-m & 100 & $R_{5}$ & 3 & 100 & 1.0\tabularnewline
\hline 
Fig. \ref{fig:2BGfree_energy_diff}d & 100 & $R_{5}$ & 3 & 100 & 0.25, 0.5, 1, 2, 3\tabularnewline
\hline 
\end{tabular}
\par\end{center}

\begin{center}
\begin{tabular}{|c||c|c||c|c|}
\cline{2-5} \cline{3-5} \cline{4-5} \cline{5-5} 
\multicolumn{1}{c|}{} & \multicolumn{2}{c||}{Fig. \ref{fig:model_systems}g-m} & \multicolumn{2}{c|}{Fig. \ref{fig:2BGfree_energy_diff}d}\tabularnewline
\hline 
iter & 200 & 500 & 200 & 100\tabularnewline
\hline 
batch & 128 & 1000 & 128 & 1000\tabularnewline
\hline 
lr & 0.01 & 0.001 & 0.01 & 0.001\tabularnewline
\hline 
$w_{ML}$ & 1 & 1 & 1 & 1\tabularnewline
\hline 
$w_{KL}$ & 0 & 1 & 0 & 1\tabularnewline
\hline 
$w_{RC}$ & 0 & 0/1$^{*}$ & 0 & 0\tabularnewline
\hline 
\end{tabular}
\par\end{center}

$*$: $w_{RC}=0$ for ``green'' results and $w_{RC}=1$ for ``orange''
results in Fig. \ref{fig:model_systems}.

\paragraph{Bistable particle dimer in a Lennard-Jones fluid}

Here we simulate two-dimensional system of a bistable particle dimer
in a dense bath of $n_{s}=36$ solvent particles with Lennard-Jones
repulsion. A similar system has been proposed in \cite{NilmeyerEtAl_PNA11_NCMC}.
The configuration vector is defined by alternating $x-$ and $y-$
positions and starting with the two dimer particles:
\[
\mathbf{x}=\left[\mathbf{x}_{1x},\mathbf{x}_{1y},\mathbf{x}_{2x},\mathbf{x}_{2y},...,\mathbf{x}_{(n_{s}+2)x},\mathbf{x}_{(n_{s}+2)y}\right].
\]
Defining the dimer distance $d=\left\Vert \mathbf{x}_{1}-\mathbf{x}_{2}\right\Vert $,
and the Heaviside step function $h$, we use the potential energy:
\begin{align*}
U(\mathbf{x}) & =k_{d}(\mathbf{x}_{1x}+\mathbf{x}_{2x})^{2}+k_{d}\mathbf{x}_{1y}^{2}+k_{d}\mathbf{x}_{2y}^{2}\\
 & +\frac{1}{4}a(d-d_{0})^{4}-\frac{1}{2}b(d-d_{0})^{2}+c(d-d_{0})^{4}\\
 & +\sum_{i=1}^{n+2}h(-\mathbf{x}_{ix}-l_{\mathrm{box}})k_{\mathrm{box}}(-\mathbf{x}_{ix}-l_{\mathrm{box}})^{2}+\sum_{i=1}^{n+2}h(\mathbf{x}_{ix}-l_{\mathrm{box}})k_{\mathrm{box}}(\mathbf{x}_{ix}-l_{\mathrm{box}})^{2}\\
 & +\sum_{i=1}^{n+2}h(-\mathbf{x}_{iy}-l_{\mathrm{box}})k_{\mathrm{box}}(-\mathbf{x}_{iy}-l_{\mathrm{box}})^{2}+\sum_{i=1}^{n+2}h(\mathbf{x}_{iy}-l_{\mathrm{box}})k_{\mathrm{box}}(\mathbf{x}_{iy}-l_{\mathrm{box}})^{2}\\
 & +\epsilon\sum_{i=1}^{n+1}\sum_{j=i+1,j\neq2}^{n+2}\left(\frac{\sigma}{\left\Vert \mathbf{x}_{i}-\mathbf{x}_{j}\right\Vert }\right)^{12}
\end{align*}
where the five rows correspond to: (1) Constraints for the center
and the $y$-position of the particle dimer, (2) particle dimer interaction,
(3,4) box constraints in $x-$ and $y-$direction, (5) particle repulsion.
The following parameter values were used (all in reduced units):
\begin{center}
\begin{tabular}{|c|c|c|c|c|c|c|c|c|c|}
\hline 
Parameter & $\epsilon$ & $\sigma$ & $k_{d}$ & $d_{0}$ & $a$ & $b$ & $c$ & $l_{\mathrm{box}}$ & $k_{\mathrm{box}}$\tabularnewline
\hline 
\hline 
Value & 1.0 & 1.1 & 20.0 & 1.5 & 25.0 & 10.0 & -0.5 & 3.0 & 100.0\tabularnewline
\hline 
\end{tabular}
\par\end{center}

To initialize training, we run Metropolis Monte Carlo simulations
with a Metropolis step length of $\sigma_{\mathrm{Metro}}=0.02\sqrt{\tau}$,
where $\tau$ is the relative temperature. To estimate the time taken
for a return-trip between open and closed dimer states, we take the
same approach as for the double-well system above: We conduct a simulation
with $10^{6}$ simulation steps for a system with maximally flattened
energy ($a=2.5$ and $b=1.0$). Still no transition from closed to
open states occur, we thus estimate the \emph{lower bound} for the
return trip to be $t=10^{6}\exp(B-B_{\mathrm{flat}})\approx1.2\cdot10^{12}$
where $B,B_{\mathrm{flat}}$ are the intrinsic barrier heights for
the unchanged and flattened system.

For validation of the free energy profiles predicted in Fig. \ref{fig:particle_dimer}e
and Fig. \ref{fig:2BGfree_energy_diff}e, we perform Umbrella Sampling
simulations \cite{Torrie_JCompPhys23_187} for each relative temperature
using 35 Umbrella potentials on the dimer distance between values
of $0.5$ and $2.5$ and with a force constant of $500$ (reduced
units). Each umbrella simulation was $50,000$ steps, and to avoid
hysteresis effects, we ran the umbrella sequence forward and backward,
resulting in a total of $3\cdot70\cdot50,000=10.5$ million simulation
steps for Fig. \ref{fig:particle_dimer}e. For Fig. \ref{fig:2BGfree_energy_diff}e
we ran 3 such simulations at each of 5 temperatures, resulting in
$52.5$ million simulation steps.

For initializing the training by example (ML), $10^{5}$ simulation
steps are stored for the ``open'' and ``closed'' dimer states,
with no transitions between these states occurring in the simulations.
For the free energy difference approach using two Boltzmann Generators
(Fig. \ref{fig:2BGfree_energy_diff}e) only $10,000$ simulation steps
were used and Gaussian noise with a standard deviation of 0.05 was
added to the configurations. In order to avoid having to learn the
permutational invariance of the diffusing solvent particles from the
data, we remove this invariance by relabeling solvent particles using
the Hungarian algorithm \cite{Kuhn_Naval55_HungarianMethod}.

Boltzmann Generator training was done using the following hyper-parameters:
\begin{center}
\begin{tabular}{|c|c|c|c|c|c|}
\hline 
Results figure & Input samples & Network & $l_{\mathrm{hidden}}$ & $n_{\mathrm{hidden}}$ & temperatures\tabularnewline
\hline 
\hline 
Fig. \ref{fig:particle_dimer} & 100,000 & $\mathrm{R}_{8}$ & 3 & 200 & 0.25, 0.5, 0.75, 1, 1.5, 2, 3, 4\tabularnewline
\hline 
Fig. \ref{fig:2BGfree_energy_diff}e & 10,000 & $WR_{8}$ & 4 & 100 & 1, 2, 3\tabularnewline
\hline 
\end{tabular}
\par\end{center}

and following training schedules:

{\footnotesize{}}%
\begin{tabular}{|c||c|c|c|c|c|c||c|c|c|c|c|c|c|c|}
\cline{2-15} \cline{3-15} \cline{4-15} \cline{5-15} \cline{6-15} \cline{7-15} \cline{8-15} \cline{9-15} \cline{10-15} \cline{11-15} \cline{12-15} \cline{13-15} \cline{14-15} \cline{15-15} 
\multicolumn{1}{c||}{} & \multicolumn{6}{c||}{{\footnotesize{}Fig. \ref{fig:particle_dimer}}} & \multicolumn{8}{c|}{{\footnotesize{}Fig. \ref{fig:2BGfree_energy_diff}e}}\tabularnewline
\hline 
{\footnotesize{}iter} & {\footnotesize{}20} & {\footnotesize{}200} & {\footnotesize{}300} & {\footnotesize{}300} & {\footnotesize{}1000} & {\footnotesize{}2000} & {\footnotesize{}100} & {\footnotesize{}40} & {\footnotesize{}40} & {\footnotesize{}40} & {\footnotesize{}40} & {\footnotesize{}40} & {\footnotesize{}100} & {\footnotesize{}200}\tabularnewline
\hline 
{\footnotesize{}batch} & {\footnotesize{}256} & {\footnotesize{}8000} & {\footnotesize{}8000} & {\footnotesize{}8000} & {\footnotesize{}8000} & {\footnotesize{}8000} & {\footnotesize{}128} & {\footnotesize{}1000} & {\footnotesize{}1000} & {\footnotesize{}1000} & {\footnotesize{}1000} & {\footnotesize{}1000} & {\footnotesize{}1000} & {\footnotesize{}1000}\tabularnewline
\hline 
{\footnotesize{}lr} & {\footnotesize{}$\mathrm{10^{-3}}$} & {\footnotesize{}$\mathrm{10^{-4}}$} & {\footnotesize{}$\mathrm{10^{-4}}$} & {\footnotesize{}$\mathrm{10^{-4}}$} & {\footnotesize{}$\mathrm{10^{-4}}$} & {\footnotesize{}$\mathrm{10^{-4}}$} & {\footnotesize{}$\mathrm{10^{-3}}$} & {\footnotesize{}$\mathrm{10^{-4}}$} & {\footnotesize{}$\mathrm{10^{-4}}$} & {\footnotesize{}$\mathrm{10^{-4}}$} & {\footnotesize{}$\mathrm{10^{-4}}$} & {\footnotesize{}$\mathrm{10^{-4}}$} & {\footnotesize{}$\mathrm{10^{-4}}$} & {\footnotesize{}$\mathrm{10^{-4}}$}\tabularnewline
\hline 
{\footnotesize{}$w_{ML}$} & {\footnotesize{}1} & {\footnotesize{}100} & {\footnotesize{}100} & {\footnotesize{}100} & {\footnotesize{}20} & {\footnotesize{}0.01} & {\footnotesize{}1} & {\footnotesize{}1000} & {\footnotesize{}300} & {\footnotesize{}100} & {\footnotesize{}50} & {\footnotesize{}20} & {\footnotesize{}5} & {\footnotesize{}1}\tabularnewline
\hline 
{\footnotesize{}$w_{KL}$} & {\footnotesize{}0} & {\footnotesize{}1} & {\footnotesize{}1} & {\footnotesize{}1} & {\footnotesize{}1} & {\footnotesize{}1} & {\footnotesize{}-} & {\footnotesize{}1} & {\footnotesize{}1} & {\footnotesize{}1} & {\footnotesize{}1} & {\footnotesize{}1} & {\footnotesize{}1} & {\footnotesize{}1}\tabularnewline
\hline 
{\footnotesize{}$w_{RC}$} & {\footnotesize{}0} & {\footnotesize{}1} & {\footnotesize{}5} & {\footnotesize{}10} & {\footnotesize{}10} & {\footnotesize{}10} & {\footnotesize{}0} & {\footnotesize{}0} & {\footnotesize{}0} & {\footnotesize{}0} & {\footnotesize{}0} & {\footnotesize{}0} & {\footnotesize{}0} & {\footnotesize{}0}\tabularnewline
\hline 
{\footnotesize{}$E_{\mathrm{high}}$} & {\footnotesize{}-} & {\footnotesize{}$10^{4}$} & {\footnotesize{}$10^{4}$} & {\footnotesize{}$10^{4}$} & {\footnotesize{}2000} & {\footnotesize{}1000} & {\footnotesize{}-} & {\footnotesize{}$10^{6}$} & {\footnotesize{}$10^{6}$} & {\footnotesize{}$10^{5}$} & {\footnotesize{}$5\cdot10^{4}$} & {\footnotesize{}$5\cdot10^{4}$} & {\footnotesize{}$5\cdot10^{4}$} & {\footnotesize{}$5\cdot10^{4}$}\tabularnewline
\hline 
\end{tabular}{\footnotesize\par}

For Fig. \ref{fig:2BGfree_energy_diff}e, we used a total of about
$10^{6}$ energy calls for training of both Boltzmann Generators and
computing their free energy differences
\begin{center}
\par\end{center}

\paragraph{Hyper-parameter optimization}

While the results shown in this paper appear to be robust over different
network architectures, we demonstrate on the particle dimer as an
example how hyper-parameter optimization can be conducted for Boltzmann
Generators, and used the resulting hyper-parameters for the results
in Fig. \ref{fig:particle_dimer}. The hyper-parameters were chosen
by minimizing the estimator variance for the free energy profile along
dimer distance $d$. Each trained network makes predictions for the
free energy profile shown in Fig. \ref{fig:particle_dimer}e. Using
bootstrapping the standard error over all free energies along the
profile between $d=[0.5,2.5]$ are computed, resulting in $(\epsilon_{0.5},\epsilon_{1.0},\epsilon_{2.0})$
for the three temperatures and $\bar{\epsilon}=\sqrt{\epsilon_{0.5}^{2}+\epsilon_{1.0}^{2}+\epsilon_{2.0}^{2}}$
as a total estimator error. Results of hyper-parameter optimization
are shown in Table S\ref{tab:hyperparameters_particle_dimer}.

\paragraph{Bovine Pancreatic Trypsin Inhibitor}

To treat more complicated molecular systems, a linking mechanism was
implemented to exchange system coordinates, potential energies and
forces in between the TensorFlow \cite{tensorflow2015} and OpenMM
software libraries \cite{EastmanEtAl_JCTC13_OpenMM}.

We set up an all-atom model of the bovine pancreatic trypsin inhibitor
(BPTI) protein which has been characterized extensively by biophysical
experiment and molecular simulations, using the published crystal
structure topology (pdb: 5PTI \cite{Wlodawer1984}) and AMBER 99 SB
ILDN \cite{LindorffLarsen2010} parameters to model intramolecular
interactions of the protein, and solvation effects were treated implicitly
using a generalized Born model (GBSA-OBC) with parameters adopted
for use for AMBER99 and it variants \cite{Onufriev2004,Ponder}.

We generated data for ML based training using conformational states
by sampling 6 initial configuration from a previously published 1
millisecond simulation of BPTI in explicit solvation \cite{Shaw_Science10_Anton}
corresponding to high density areas in the slow collective coordinates
\cite{NoeClementi_COSB17_SlowCVs}. The six configurations were chosen
using $k$-means clustering of the 15 leading components computed
with TICA (time-lagged independent component analysis) \cite{PerezEtAl_JCP13_TICA}
at 1 $\mu s$ lag-time and using the following features: cosines and
sines of all backbone torsions and $\chi_{1}$ angles of cysteine
residues forming a flexible disulfide bond (residue 14 and 35). The
selected frames correspond to time-points: 46.396, 92.682, 70.339
87.889, 827.930 and 831.050 microseconds in the original trajectory.
From each of these six frames, 20 nanoseconds of MD simulation was
conducted at temperature 300 K using the forcefield as specified above
using a Langevin integration approach with an integration time-step
of 2 femtoseconds. Configurations were stored every $0.2$ picoseconds.

For BPTI we used Boltzmann Generators with a mixed coordinate transformation
layer $M$ as first layer. The Cartesian set consisted of heavy backbone
atoms (N, C$\alpha$, C') and the heavy side-chain atoms of disulfide
bridges. The rest of the atoms ($H_{\alpha}$, backbone $O$ and side-chains
not involved in disulfide bridges) defined the internal coordinate
set. For all BPTI Boltzmann Generators, the MD data was subsampled
to 100,000 configurations, starting from 6 MD datasets in Figs. \ref{fig:bpti}
and 2 MD datasets in Fig. \ref{fig:2BGfree_energy_diff}f. The hyper-parameters
were chosen as follows:
\begin{center}
\begin{tabular}{|c|c|c|c|c|}
\hline 
Results figure & Network & $l_{\mathrm{hidden}}$ & $n_{\mathrm{hidden}}$ & temperatures\tabularnewline
\hline 
\hline 
Fig. \ref{fig:bpti} & $MR_{8}$ & 4 & 256, 128, 256 & 1\tabularnewline
\hline 
Fig. \ref{fig:2BGfree_energy_diff}e & $MR_{8}$ & 4 & 200, 100, 200 & 0.9 0.95 1.0 1.05 1.1 1.15 1.2\tabularnewline
\hline 
\end{tabular}
\par\end{center}

For the results in Fig. \ref{fig:bpti}, we made use of a two dimensional
reaction coordinate loss defined by the two first time-lagged independent
components estimated using a previously published molecular dynamics
trajectory. We used the sines and cosines of backbone torsion angles
and side-chain $\chi_{1}$ angles of Cys 14 and Cys 38 as basis functions
and estimated the projection with a with a lag-time of 100 nanoseconds.
The loss is defined as the negative entropy of a batch distribution
projected onto these coordinates, estimated using a soft binning with
a $11\times11$ grid spanning the values $[-1,6]$ and $[-2.1,1.75]$
respectively. 

Training was initiated by three times 2000 iterations of maximum likelihood
with batch sizes 128, 256 and 512, respectively. Subsequently, following
stages of mixed maximum likelihood and energy-based training was conducted
for the Boltzmann Generator in Fig. \ref{fig:bpti}, where \foreignlanguage{english}{$w_{RC}=20.0$,
$w_{\text{torsion}}=1.0$, $w_{\text{ML}}=1.0$} were used throughout:
\begin{center}
{\footnotesize{}}%
\begin{tabular}{|c||c|c|c|c|c|c|c|c|c|c|c|c|}
\hline 
{\footnotesize{}iter} & {\footnotesize{}15} & {\footnotesize{}15} & {\footnotesize{}15} & {\footnotesize{}15} & {\footnotesize{}15} & {\footnotesize{}15} & {\footnotesize{}20} & {\footnotesize{}20} & {\footnotesize{}30} & {\footnotesize{}50} & {\footnotesize{}50} & {\footnotesize{}300}\tabularnewline
\hline 
{\footnotesize{}batch} & {\footnotesize{}5000} & {\footnotesize{}5000} & {\footnotesize{}5000} & {\footnotesize{}5000} & {\footnotesize{}5000} & {\footnotesize{}5000} & {\footnotesize{}5000} & {\footnotesize{}5000} & {\footnotesize{}5000} & {\footnotesize{}5000} & {\footnotesize{}5000} & {\footnotesize{}5000}\tabularnewline
\hline 
{\footnotesize{}lr} & {\footnotesize{}$10^{-4}$} & {\footnotesize{}$10^{-4}$} & {\footnotesize{}$10^{-4}$} & {\footnotesize{}$10^{-4}$} & {\footnotesize{}$10^{-4}$} & {\footnotesize{}$10^{-4}$} & {\footnotesize{}$10^{-4}$} & {\footnotesize{}$10^{-4}$} & {\footnotesize{}$10^{-4}$} & {\footnotesize{}$10^{-4}$} & {\footnotesize{}$10^{-4}$} & {\footnotesize{}$10^{-4}$}\tabularnewline
\hline 
{\footnotesize{}$E_{\mathrm{high}}$} & {\footnotesize{}$10^{10}$} & {\footnotesize{}$10^{9}$} & {\footnotesize{}$10^{8}$} & {\footnotesize{}$10^{7}$} & {\footnotesize{}$10^{6}$} & {\footnotesize{}$10^{5}$} & {\footnotesize{}$10^{5}$} & {\footnotesize{}$10^{5}$} & {\footnotesize{}$10^{4}$} & {\footnotesize{}$10^{4}$} & {\footnotesize{}$10^{3}$} & {\footnotesize{}$10^{3}$}\tabularnewline
\hline 
{\footnotesize{}$w_{KL}$} & {\footnotesize{}$10^{-12}$} & {\footnotesize{}$10^{-6}$} & {\footnotesize{}$10^{-5}$} & {\footnotesize{}$10^{-4}$} & {\footnotesize{}$10^{-3}$} & {\footnotesize{}$10^{-3}$} & {\footnotesize{}$5\cdot10^{-3}$} & {\footnotesize{}$10^{-3}$} & {\footnotesize{}$5\cdot10^{-3}$} & {\footnotesize{}$5\cdot10^{-2}$} & {\footnotesize{}$5\cdot10^{-2}$} & {\footnotesize{}$5\cdot10^{-2}$}\tabularnewline
\hline 
\end{tabular}{\footnotesize\par}
\par\end{center}

For the computation of free energy differences using two Boltzmann
Generators we used $20$ ns MD simulation from states ``X'' and
``O'' as shown in Fig. \ref{fig:2BGfree_energy_diff}f, and the
following simplified training protocol:
\begin{center}
\begin{tabular}{|c||c|c|c|c|c|c|c|c|c|c|}
\hline 
{\footnotesize{}iter} & {\footnotesize{}2000} & {\footnotesize{}2000} & {\footnotesize{}2000} & {\footnotesize{}30} & {\footnotesize{}30} & {\footnotesize{}30} & {\footnotesize{}30} & {\footnotesize{}30} & {\footnotesize{}30} & {\footnotesize{}400}\tabularnewline
\hline 
{\footnotesize{}batch} & {\footnotesize{}128} & {\footnotesize{}256} & {\footnotesize{}512} & {\footnotesize{}5000} & {\footnotesize{}5000} & {\footnotesize{}5000} & {\footnotesize{}5000} & {\footnotesize{}5000} & {\footnotesize{}5000} & {\footnotesize{}5000}\tabularnewline
\hline 
{\footnotesize{}$E_{\mathrm{high}}$} & - & - & - & {\footnotesize{}$10^{10}$} & {\footnotesize{}$10^{9}$} & {\footnotesize{}$10^{8}$} & {\footnotesize{}$10^{7}$} & {\footnotesize{}$10^{6}$} & {\footnotesize{}$10^{5}$} & {\footnotesize{}$10^{4}$}\tabularnewline
\hline 
{\footnotesize{}lr} & {\footnotesize{}$10^{-3}$} & {\footnotesize{}$10^{-3}$} & {\footnotesize{}$10^{-3}$} & {\footnotesize{}$10^{-4}$} & {\footnotesize{}$10^{-4}$} & {\footnotesize{}$10^{-4}$} & {\footnotesize{}$10^{-4}$} & {\footnotesize{}$10^{-4}$} & {\footnotesize{}$10^{-4}$} & {\footnotesize{}$10^{-4}$}\tabularnewline
\hline 
{\footnotesize{}$w_{KL}$} & {\footnotesize{}0} & {\footnotesize{}0} & {\footnotesize{}0} & {\footnotesize{}$10^{-7}$} & {\footnotesize{}$10^{-6}$} & {\footnotesize{}$10^{-5}$} & {\footnotesize{}$10^{-4}$} & {\footnotesize{}$10^{-3}$} & {\footnotesize{}$10^{-2}$} & {\footnotesize{}$10^{-2}$}\tabularnewline
\hline 
{\footnotesize{}$w_{\mathrm{tor}}$} & {\footnotesize{}-} & {\footnotesize{}-} & {\footnotesize{}-} & {\footnotesize{}0.01} & {\footnotesize{}0.1} & {\footnotesize{}0.1} & {\footnotesize{}0.1} & {\footnotesize{}1} & {\footnotesize{}1} & {\footnotesize{}1}\tabularnewline
\hline 
\end{tabular}
\par\end{center}

\clearpage

\subsection*{Supplementary Figures}

\renewcommand{\thefigure}{S\arabic{figure}}
\setcounter{figure}{0} 
\begin{center}
\begin{figure}[H]
\begin{centering}
\includegraphics[width=1\textwidth]{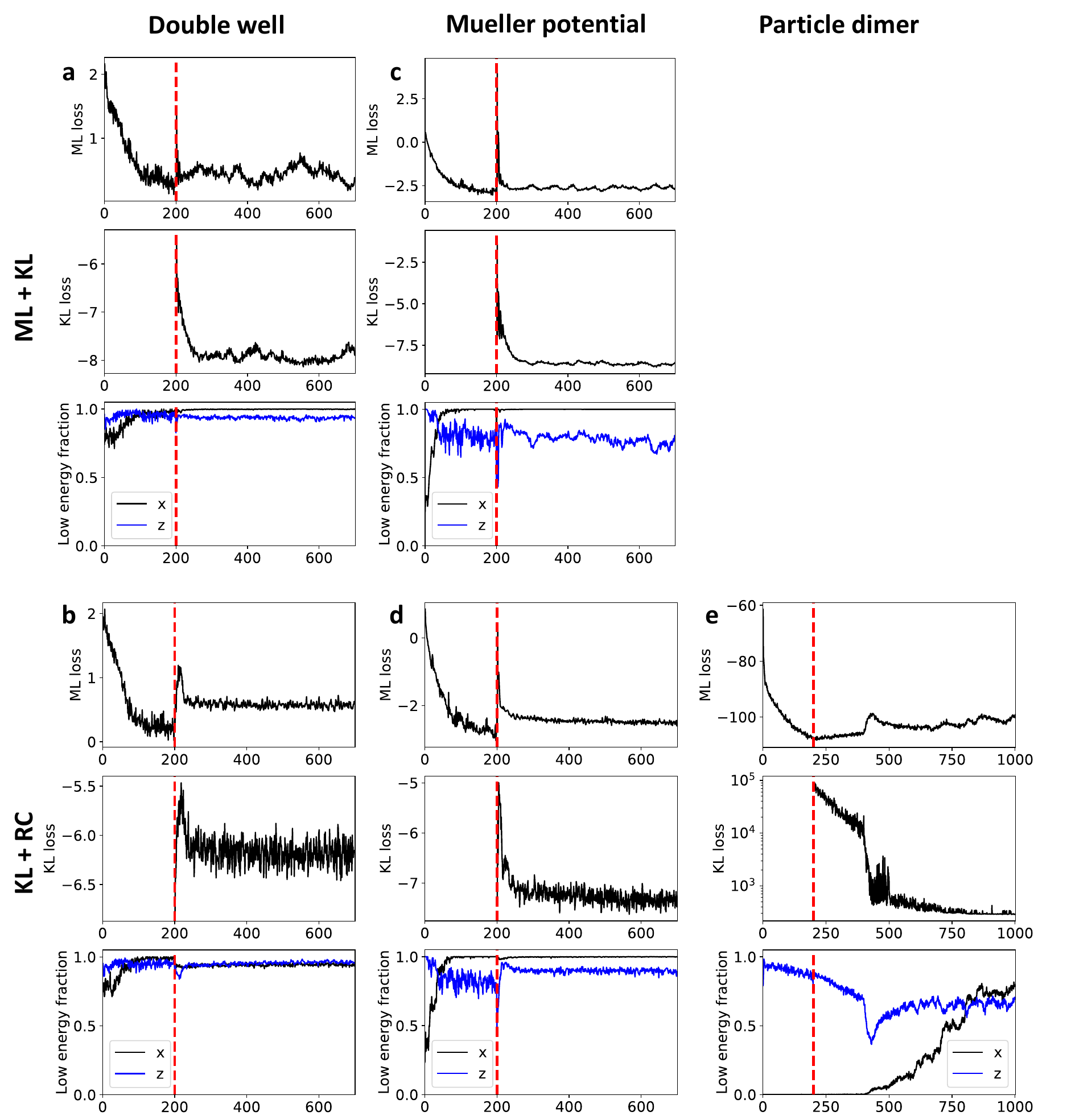}
\par\end{centering}
\caption{\label{fig_convergence1}\textbf{Convergence metrics for Boltzmann
Generators reported in Fig. \ref{fig:model_systems} and Fig. \ref{fig:particle_dimer}}.
Boltzmann Generators were trained with by energy and example (top
row), or by energy and using a reaction coordinate loss (bottom row).
Panels show: ML loss $J_{ML}$ (top), KL loss $J_{KL}$ (middle) and
the low-energy fractions in configuration space $\mathbf{x}$ and
in latent space $\mathbf{z}$ (bottom). These are defined by the fraction
of Boltzmann Generator samples ($\mathbf{z}\rightarrow\mathbf{x}$)
whose configuration energies are within the 99\% percentile of the
energies of the input data, and the fraction of input data that when
mapped to latent space ($\mathbf{x}\rightarrow\mathbf{z}$) are within
the 99\% percentile of the energy distribution of a harmonic oscillator
with the respective dimension.\textbf{ a}-\textbf{b}): Double well
potential. \textbf{c}-\textbf{d}): Mueller potential. \textbf{e})
Solvated particle dimer.}
\end{figure}
\par\end{center}

\begin{center}
\begin{figure}[H]
\begin{centering}
\includegraphics[width=1\textwidth]{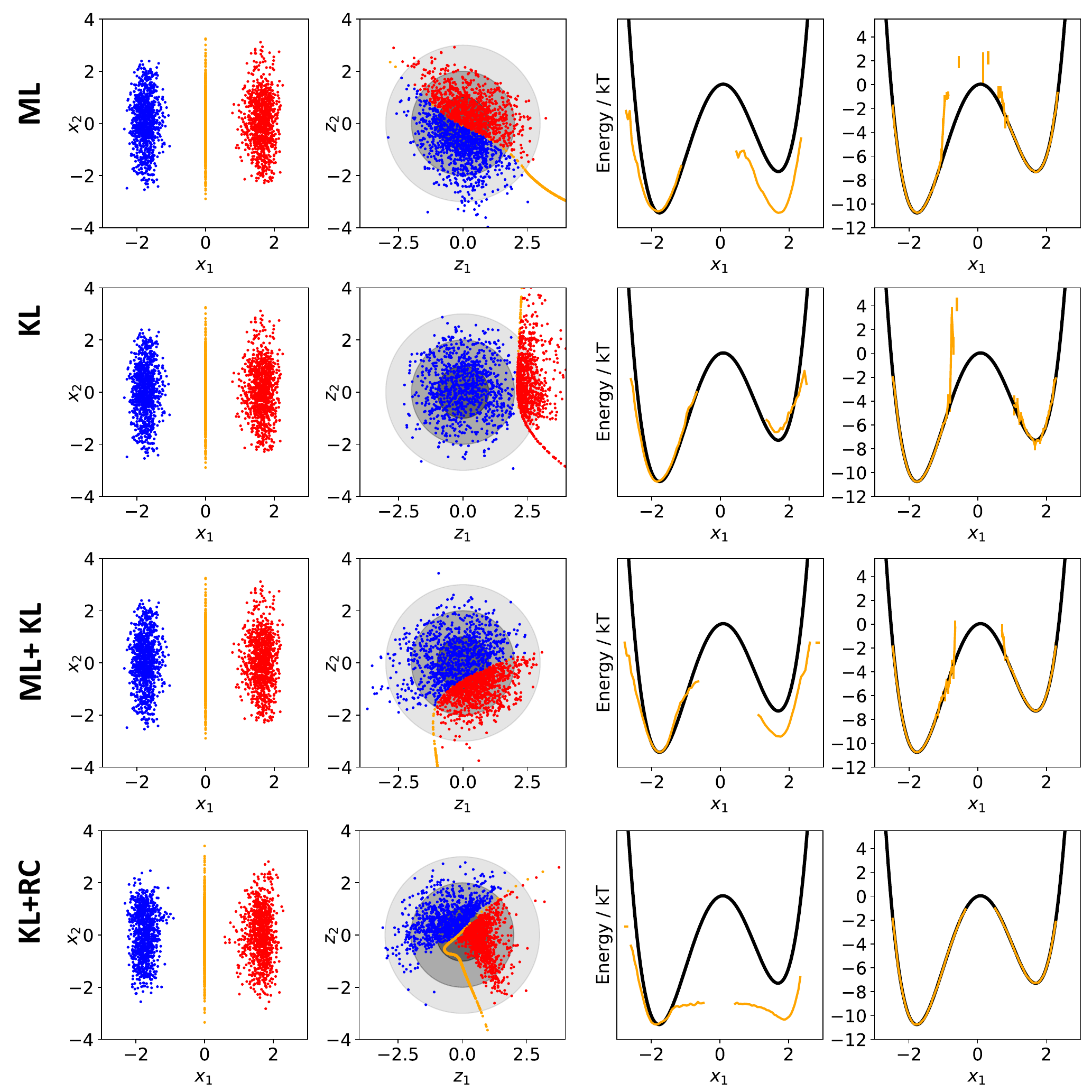}
\par\end{centering}
\caption{\label{fig_training_methods_RealNVP}\textbf{Different training methods
for Boltzmann Generators using the double well example (Fig. \ref{fig:model_systems})}.
Columns show: (1) distribution in configuration space $\mathbf{x}$,
(2) distribution in latent space $\mathbf{z}$, (3) free energy of
Boltzmann Generator output $p_{X}(\mathbf{x})$ along $x_{1}$, (4)
free energy after reweighting, vertical bars show uncertainties (one
standard deviation, 68\% percentile). Training proceeds by 200 iterations
of ML and then 500 iterations of the method given in the rows, using
equal weights for these modes. Training by example (ML) only reproduces
the distribution of the training data, which can be reweighted to
the Boltzmann distribution in this low-dimensional example but reweighting
from the ML-generated distribution fails for high-dimensional examples.
Training by energy (KL) alone tends to collapse to a single metastable
state. ML+KL combined samples closer to the Boltzmann distribution
than ML and avoids metastable state collapse, but samples high-energy
transition states with low probability. KL+RC performs best in this
example.}
\end{figure}
\begin{figure}
\begin{centering}
\includegraphics[width=0.6\textwidth]{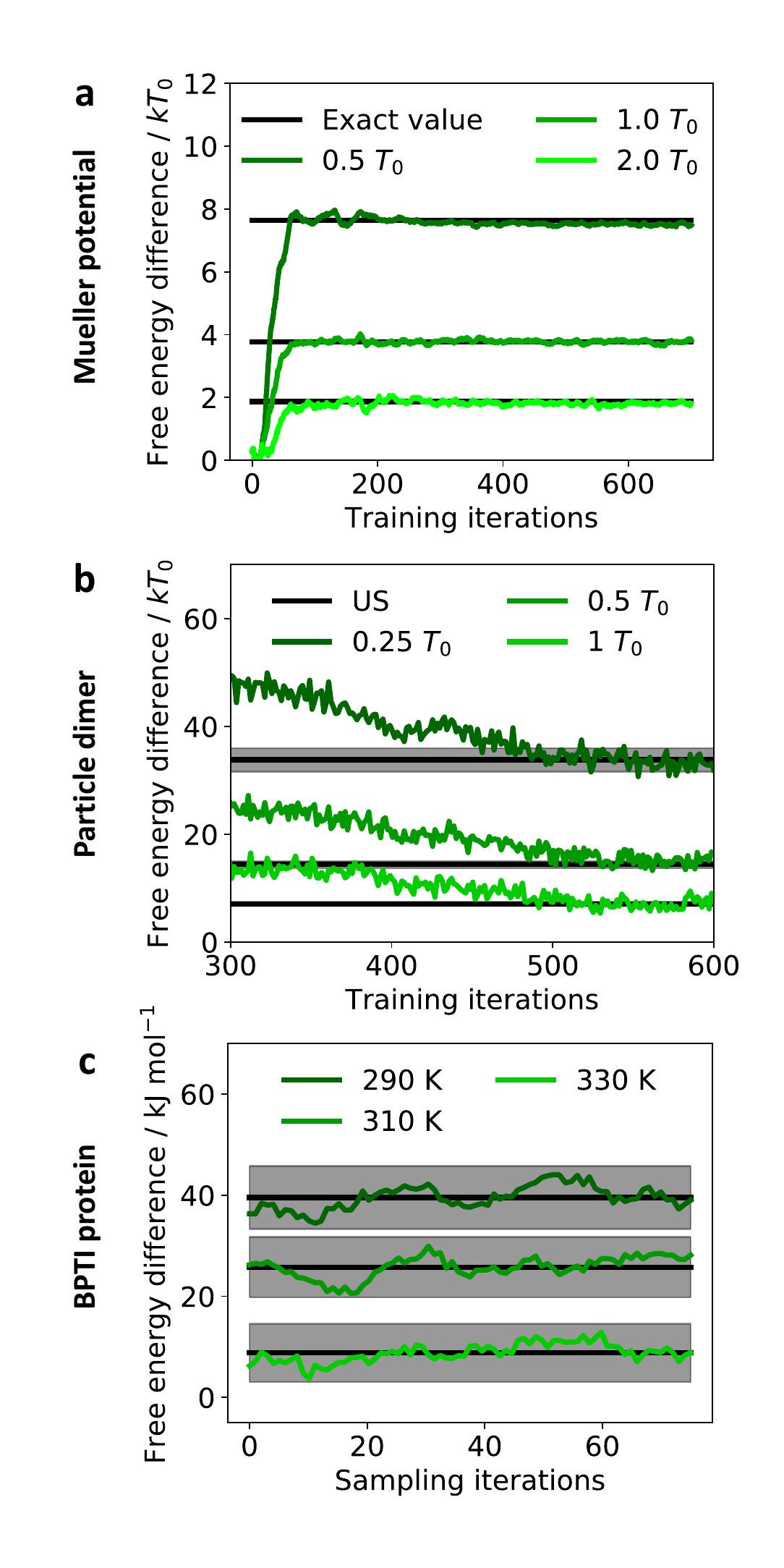}
\par\end{centering}
\caption{\label{fig_convergence_2BG}\textbf{Convergence of free energy differences
shown in Fig. \ref{fig:2BGfree_energy_diff}}. \textbf{a}) Mueller
potential. \textbf{b}) Solvated Particle dimer. Convergence is shown
for the last two training cycles, as the free energies are far from
converged in earlier stages. References are from umbrella sampling,
with intervals showing one standard error from 3 repeats. \textbf{c})
BPTI protein. For better performance, the free energy differences
were not estimated during the training schedule reported in the Supp.
Mat. Instead 75 training iterations were appended afterwards, and
in each iteration 1000 samples of $J_{KL}$ were obtained at each
temperature for free energy estimation. Black lines / grey intervals
show the mean and one standard error estimates from this phase using
5 independent repeats.}
\end{figure}
\par\end{center}

\renewcommand{\thetable}{S\arabic{table}}
\setcounter{table}{0} 
\begin{center}
\begin{table}[H]
\begin{centering}
\begin{tabular}{|c|c|c|c|c|c|c|c|c|}
\hline 
Architecture & $nl_{\mathrm{layers}}$ & $nl_{hidden}$ & $w_{ML}$ & $w_{RC}$ & $\epsilon_{0.5}$ & $\epsilon_{1}$ & $\epsilon_{2}$ & $\sqrt{\sum\epsilon^{2}}$\tabularnewline
\hline 
$\mathrm{R}_{8}$ & 4 & 200 & 0.1 & 10.0 & 1.62 & 2.07 & 2.04 & 3.33\tabularnewline
\hline 
$\mathrm{R}_{4}$ & $\cdot$ & $\cdot$ & $\cdot$ & $\cdot$ & 2.23 & 1.83 & 1.53 & 3.27\tabularnewline
\hline 
$\mathrm{R}_{6}$ & $\cdot$ & $\cdot$ & $\cdot$ & $\cdot$ & 1.69 & 1.64 & 2.29 & 3.28\tabularnewline
\hline 
$\mathrm{R}_{12}$ & $\cdot$ & $\cdot$ & $\cdot$ & $\cdot$ & 1.49 & 1.85 & 2.0 & 3.10\tabularnewline
\hline 
\multirow{1}{*}{$\mathrm{R}_{8}$} & 3 & 200 & 0.1 & 10.0 & 1.51 & 1.97 & 1.64 & 2.97\tabularnewline
\hline 
$\mathrm{R}_{4}$ & $\cdot$ & $\cdot$ & $\cdot$ & $\cdot$ & 1.41 & 1.59 & 1.78 & 2.77\tabularnewline
\hline 
$\mathrm{R}_{6}$ & $\cdot$ & $\cdot$ & $\cdot$ & $\cdot$ & 1.49 & 1.73 & 1.76 & 2.88\tabularnewline
\hline 
$\mathrm{R}_{12}$ & $\cdot$ & $\cdot$ & $\cdot$ & $\cdot$ & 1.84 & 1.28 & 2.24 & 3.17\tabularnewline
\hline 
$\mathrm{R}_{8}$ & 2 & $\cdot$ & $\cdot$ & $\cdot$ & 1.85 & 1.58 & 2.50 & 3.48\tabularnewline
\hline 
\textbf{$\cdot$} & 4 & $\cdot$ & $\cdot$ & $\cdot$ & 1.69 & 1.51 & 1.52 & 2.73\tabularnewline
\hline 
$\cdot$ & 3 & 50 & $\cdot$ & $\cdot$ & 1.32 & 1.71 & 2.11 & 3.02\tabularnewline
\hline 
$\cdot$ & $\cdot$ & 100 & $\cdot$ & $\cdot$ & 2.85 & 2.05 & 2.16 & 4.12\tabularnewline
\hline 
\textbf{$\cdot$} & \textbf{$\cdot$} & \textbf{200} & \textbf{0.01} & \textbf{$\cdot$} & \textbf{1.58} & \textbf{1.33} & \textbf{1.33} & \textbf{2.45}\tabularnewline
\hline 
$\cdot$ & $\cdot$ & $\cdot$ & 1.0 & $\cdot$ & 1.87 & 1.93 & 1.63 & 3.15\tabularnewline
\hline 
$\cdot$ & $\cdot$ & $\cdot$ & 0.1 & 1.0 & 1.66 & 1.83 & 1.75 & 3.02\tabularnewline
\hline 
$\cdot$ & $\cdot$ & $\cdot$ & $\cdot$ & 5.0 & 1.73 & 1.72 & 1.81 & 3.03\tabularnewline
\hline 
$\cdot$ & $\cdot$ & $\cdot$ & $\cdot$ & 20.0 & 1.88 & 2.06 & 1.84 & 3.34\tabularnewline
\hline 
\end{tabular}
\par\end{centering}
\caption{\label{tab:hyperparameters_particle_dimer}\textbf{Hyper-parameter
selection for the particle dimer}. In the architecture, $R$ corresponds
to a RealNVP block, i.e. two layers with channel swaps (Fig. \ref{fig:illustration}b).
The subscript indicates the number of repetitions, e.g. $\mathrm{R}_{4}=\mathrm{RRRR}$,
corresponding to eight single layers. All nonlinear transformations
($T$, $S$) the given number of layers ($nl_{layers}$) and hidden
nodes ($nl_{hidden}$). All networks were trained on the following
range of relative temperatures: $\tau\in[0.1,0.25,0.5,0.75,1,1.5,2,3,4]$
and used $w_{KL}=1.0$.}
\end{table}
\par\end{center}

\begin{thebibliography}{10}

\bibitem{Torrie_JCompPhys23_187}
G.~M. Torrie, J.~P. Valleau, {\it J. Comp. Phys.\/} {\bf 23}, 187 (1977).

\bibitem{Grubmueller_PhysRevE52_2893}
H.~Grubm\"{u}ller, {\it Phys. Rev. E\/} {\bf 52}, 2893 (1995).

\bibitem{LaioParrinello_PNAS99_12562}
A.~Laio, M.~Parrinello, {\it Proc. Natl. Acad. Sci. USA\/} {\bf 99}, 12562
  (2002).

\bibitem{HeninEtAl_JCTC10_ABF}
J.~H{\'e}nin, G.~Fiorin, C.~Chipot, M.~L. Klein, {\it J. Chem. Theory
  Comput.\/} {\bf 6}, 35 (2010).

\bibitem{SwendsenWang_PRL86_ParallelTempering}
R.~H. S. J.~S. Wang, {\it Phys. Rev. Lett.\/} {\bf 57}, 2607 (1986).

\bibitem{Hukushima_JPSJapan65_1604}
K.~Hukushima, K.~Nemoto, {\it J. Phys. Soc. (Jap.)\/} {\bf 65}, 1604 (1996).

\bibitem{MarinariParisi_EPL92_SimulatedTempering}
E.~Marinari, G.~Parisi, {\it Europhy. Lett.\/} {\bf 19}, 451 (1992).

\bibitem{Kirkwood_JCP35_CouplingParameterMethod}
J.~G. Kirkwood, {\it J. Chem. Phys.\/} {\bf 3}, 300 (1935).

\bibitem{FrenkelSmit_MolecularSimulation}
B.~S. Daan~Frenkel, {\it Understanding molecular simulation\/} (Academic Press,
  2001).

\bibitem{KlimovichShirtsMobley_JCAMD15_FreeEnergy}
P.~V. Klimovich, M.~R. Shirts, D.~L. Mobley, {\it J. Comput. Aided. Mol.
  Des.\/} {\bf 29}, 397 (2015).

\bibitem{LeCunBengioHinton_DeepLearning_Nature05}
Y.~LeCun, Y.~Bengio, G.~Hinton, {\it Nature\/} {\bf 521}, 436 (2015).

\bibitem{ZhuEtAl_PRL02_VariableTransformation}
Z.~Zhu, M.~E. Tuckerman, S.~O. Samuelson, G.~J. Martyna, {\it Phys. Rev.
  Lett.\/} {\bf 88}, 100201 (2002).

\bibitem{GoodfellowEtAl_GANs}
I.~Goodfellow, {\it et~al.\/}, {\it NIPS'14 Proceedings of the 27th
  International Conference on Neural Information Processing Systems,
  arXiv:1406.2661\/} (2014).

\bibitem{KingmaWelling_ICLR14_VAE}
D.~P. Kingma, M.~Welling, {\it Proceedings of the 2nd International Conference
  on Learning Representations (ICLR), arXiv:1312.6114\/} (2014).

\bibitem{KarrasEtAl_ProgressiveGrowingGANs}
T.~Karras, T.~Aila, S.~Laine, J.~Lehtinen, {\it Proceedings of the 7nd
  International Conference on Learning Representations (ICLR),
  arXiv:1710.10196\/}  (2018).

\bibitem{VanDenOord_WaveNet2}
A.~van~den Oord, {\it et~al.\/}, {\it 35th International Conference on Machine
  Learning (ICML), arXiv:1711.10433\/}  (2018).

\bibitem{GomezBombarelli_ACSCentral_AutomaticDesignVAE}
R.~G{\'o}mez-Bombarelli, {\it et~al.\/}, {\it ACS Cent. Sci.\/} {\bf 4}, 268
  (2018).

\bibitem{TabakVandenEijnden_CMS10_DensityEstimation}
E.~G. Tabak, E.~Vanden-Eijnden, {\it Commun. Math. Sci.\/} {\bf 8}, 217 (2010).

\bibitem{DinhDruegerBengio_NICE2015}
L.~Dinh, D.~Krueger, Y.~Bengio, {\it arXiv:1410.8516\/}  (2015).

\bibitem{DinhBengio_RealNVP}
S.~B. L.~Dinh, J. Sohl-Dickstein, {\it arXiv:1605.08803\/}  (2016).

\bibitem{RezendeEtAl_NormalizingFlows}
D.~J. Rezende, S.~Mohamed, {\it arXiv:1505.05770\/}  (2015).

\bibitem{KingmaDhariwal_NIPS18_Glow}
D.~P. Kingma, P.~Dhariwal, {\it NIPS'18 Proceedings of the 31th International
  Conference on Neural Information Processing Systems, arXiv:1807.03039\/}
  (2018).

\bibitem{GrathwohlEtAl_FFJORD}
W.~Grathwohl, R.~T.~Q. Chen, J.~Bettencourt, I.~Sutskever, D.~Duvenaud, {\it
  arXiv:1810.01367\/}  (2018).

\bibitem{BolhuisChandlerDellagoGeissler_AnnuRevPhysChem02_TPS}
P.~G. Bolhuis, D.~Chandler, C.~Dellago, P.~L. Geissler, {\it Annu. Rev. Phys.
  Chem.\/} {\bf 53}, 291 (2002).

\bibitem{NilmeyerEtAl_PNA11_NCMC}
J.~P. Nilmeier, G.~E. Crooks, D.~D.~L. Minh, J.~D. Chodera, {\it Proc. Natl.
  Acad. Sci. USA\/} {\bf 108}, E1009 (2011).

\bibitem{EastmanEtAl_JCTC13_OpenMM}
P.~Eastman, {\it et~al.\/}, {\it J. Chem. Theory Comput.\/} {\bf 9}, 461
  (2013).

\bibitem{tensorflow2015}
M.~Abadi, {\it et~al.\/}, Tensorflow: Large-scale machine learning on
  heterogeneous systems, http://tensorflow.org/ (2015).

\bibitem{Shaw_Science10_Anton}
D.~E. Shaw, {\it et~al.\/}, {\it Science\/} {\bf 330}, 341 (2010).

\bibitem{PerezEtAl_JCP13_TICA}
G.~Perez-Hernandez, F.~Paul, T.~Giorgino, G.~{D Fabritiis}, F.~No{\'e}, {\it J.
  Chem. Phys.\/} {\bf 139}, 015102 (2013).

\bibitem{SchererEtAl_JCTC15_EMMA2}
M.~K. Scherer, {\it et~al.\/}, {\it J. Chem. Theory Comput.\/} {\bf 11}, 5525
  (2015).

\bibitem{Grey2003}
M.~J. Grey, C.~Wang, A.~G. Palmer, {\it J. Am. Chem. Soc.\/} {\bf 125}, 14324
  (2003).

\bibitem{FrenkelLadd_JCP84_FreeEnergySolids}
D.~Frenkel, A.~J.~C. Ladd, {\it J. Chem. Phys.\/} {\bf 81}, 3188 (1984).

\bibitem{HooverRee_JCP68_MeltingTransition}
W.~G. Hoover, F.~H. Ree, {\it J. Chem. Phys.\/} {\bf 49}, 3609 (1968).

\bibitem{YtrebergaZuckerman_JCP06_FreeEnergies}
F.~M. Ytreberga, D.~M. Zuckerman, {\it J. Chem. Phys.\/} {\bf 124}, 104105
  (2006).

\bibitem{ChenEtAl_PNAS15_FreeEnergyLearning}
M.~Chen, T.-Q. Yu, M.~E. Tuckerman, {\it Proc. Natl. Acad. Sci. USA\/} {\bf
  112}, 3235 (2015).

\bibitem{BehlerParrinello_PRL07_NeuralNetwork}
J.~Behler, M.~Parrinello, {\it Phys. Rev. Lett.\/} {\bf 98}, 146401 (2007).

\bibitem{RuppEtAl_PRL12_QML}
M.~Rupp, A.~Tkatchenko, K.-R. M{\"u}ller, O.~A.~V. Lilienfeld, {\it Phys. Rev.
  Lett.\/} {\bf 108}, 058301 (2012).

\bibitem{ShirtsChodera_JCP08_MBAR}
M.~R. Shirts, J.~D. Chodera, {\it J. Chem. Phys.\/} {\bf 129}, 124105 (2008).

\bibitem{KingmaBa_ADAM}
D.~P. Kingma, J.~Ba, {\it Proceedings of the 4th International Conference on
  Learning Repre- sentations (ICLR), arXiv:1412.6980\/}  (2015).

\bibitem{Kuhn_Naval55_HungarianMethod}
H.~W. Kuhn, {\it Nav. Res. Logist. Quart.\/} {\bf 2}, 83 (1955).

\bibitem{Wlodawer1984}
A.~Wlodawer, J.~Walter, R.~Huber, L.~Sj\"{o}lin, {\it J. Mol. Biol.\/} {\bf
  180}, 301 (1984).

\bibitem{LindorffLarsen2010}
K.~Lindorff-Larsen, {\it et~al.\/}, {\it Proteins\/} {\bf 78}, 1950 (2010).

\bibitem{Onufriev2004}
A.~Onufriev, D.~Bashford, D.~A. Case, {\it Proteins: Structure, Function, and
  Bioinformatics\/} {\bf 55}, 383 (2004).

\bibitem{Ponder}
J.~W. Ponder, Tinker: Software tools for molecular design (2004).

\bibitem{NoeClementi_COSB17_SlowCVs}
F.~No{\'e}, C.~Clementi, {\it Curr. Opin. Struc. Biol.\/} {\bf 43}, 141 (2017).

\end{thebibliography}
\end{document}